%% file: main.tex
\documentclass[10pt,twocolumn,letterpaper]{article}

\usepackage{iccv}
\usepackage{times}
\usepackage{epsfig}
\usepackage{graphicx}
\usepackage[fleqn]{amsmath}
\usepackage{amssymb}

\usepackage[pagebackref=true,breaklinks=true,letterpaper=true,colorlinks,bookmarks=false]{hyperref}

\usepackage{multirow}
\usepackage{booktabs}
\usepackage{xcolor}
\usepackage{colortbl}
\usepackage{sidecap}
\definecolor{mygray}{gray}{.9}
\newcommand{\vect}[1]{\mbox{\boldmath$#1$}}
\iccvfinalcopy 
\usepackage[accsupp]{axessibility}  

\ificcvfinal\fi

\begin{document}

\title{FLIP: Cross-domain Face Anti-spoofing with Language Guidance}

\author{Koushik Srivatsan \quad Muzammal Naseer \quad  Karthik Nandakumar\\
Mohamed Bin Zayed University of Artificial Intelligence (MBZUAI)\\
Abu Dhabi, United Arab Emirates\\
{\tt\small \{koushik.srivatsan,  muzammal.naseer,  karthik.nandakumar\}@mbzuai.ac.ae}
}

\maketitle
\ificcvfinal\thispagestyle{empty}\fi

\begin{abstract}
    \noindent Face anti-spoofing (FAS) or presentation attack detection is an essential component of face recognition systems deployed in security-critical applications. Existing FAS methods have poor generalizability to unseen spoof types, camera sensors, and environmental conditions. Recently, vision transformer (ViT) models have been shown to be effective for the FAS task due to their ability to capture long-range dependencies among image patches. However, adaptive modules or auxiliary loss functions are often required to adapt pre-trained ViT weights learned on large-scale datasets such as ImageNet. In this work, we first show that initializing ViTs with multimodal (e.g., CLIP) pre-trained weights improves generalizability for the FAS task, which is in line with the zero-shot transfer capabilities of vision-language pre-trained (VLP) models. We then propose a novel approach for robust cross-domain FAS by grounding visual representations with the help of natural language. Specifically, we show that aligning the image representation with an ensemble of class descriptions (based on natural language semantics) improves FAS generalizability in low-data regimes. Finally, we propose a multimodal contrastive learning strategy to boost feature generalization further and bridge the gap between source and target domains. Extensive experiments on three standard protocols demonstrate that our method significantly outperforms the state-of-the-art methods, achieving better zero-shot transfer performance than five-shot transfer of ``adaptive ViTs''. Code: \url{https://github.com/koushiksrivats/FLIP}
\end{abstract}

\section{Introduction}
\label{sec:intro}

From personal devices to airport boarding gates, face recognition systems have become a ubiquitous tool for recognizing people. This may be attributed to recent advances in face recognition technology based on deep learning, as well as its simplicity and non-contact nature. However, these systems are vulnerable to face presentation attacks, where an attacker tries to spoof the identity of a bonafide individual with the help of presentation attack instruments (PAI) such as printed photos, replayed videos, or 3D synthetics masks \cite{yu2022deep}. Therefore, face anti-spoofing (FAS) or face presentation attack detection (FPAD) is essential to secure face recognition systems against presentation attacks.

Prior works \cite{zhang2020face, liu2020disentangling, yu2020face, wang2022disentangled, yu2020searching, yu2020fas, wang2022patchnet} have shown that impressive FAS accuracy can be achieved in intra-domain scenarios, where the training and test distributions are similar. However, existing FAS methods fail to generalize well to the unseen target domains due to two main reasons: (a) variations due to camera sensors, presentation attack instruments, illumination changes, and image resolution cause a large domain gap between the source and target distributions that is inherently hard to bridge; and (b) commonly used FAS benchmark datasets have limited training data, causing the model to overfit to the source domain(s). Consequently, achieving robust cross-domain FAS performance has remained an elusive challenge thus far.

\begin{figure}
    \centering
    \includegraphics[width=0.48\textwidth]{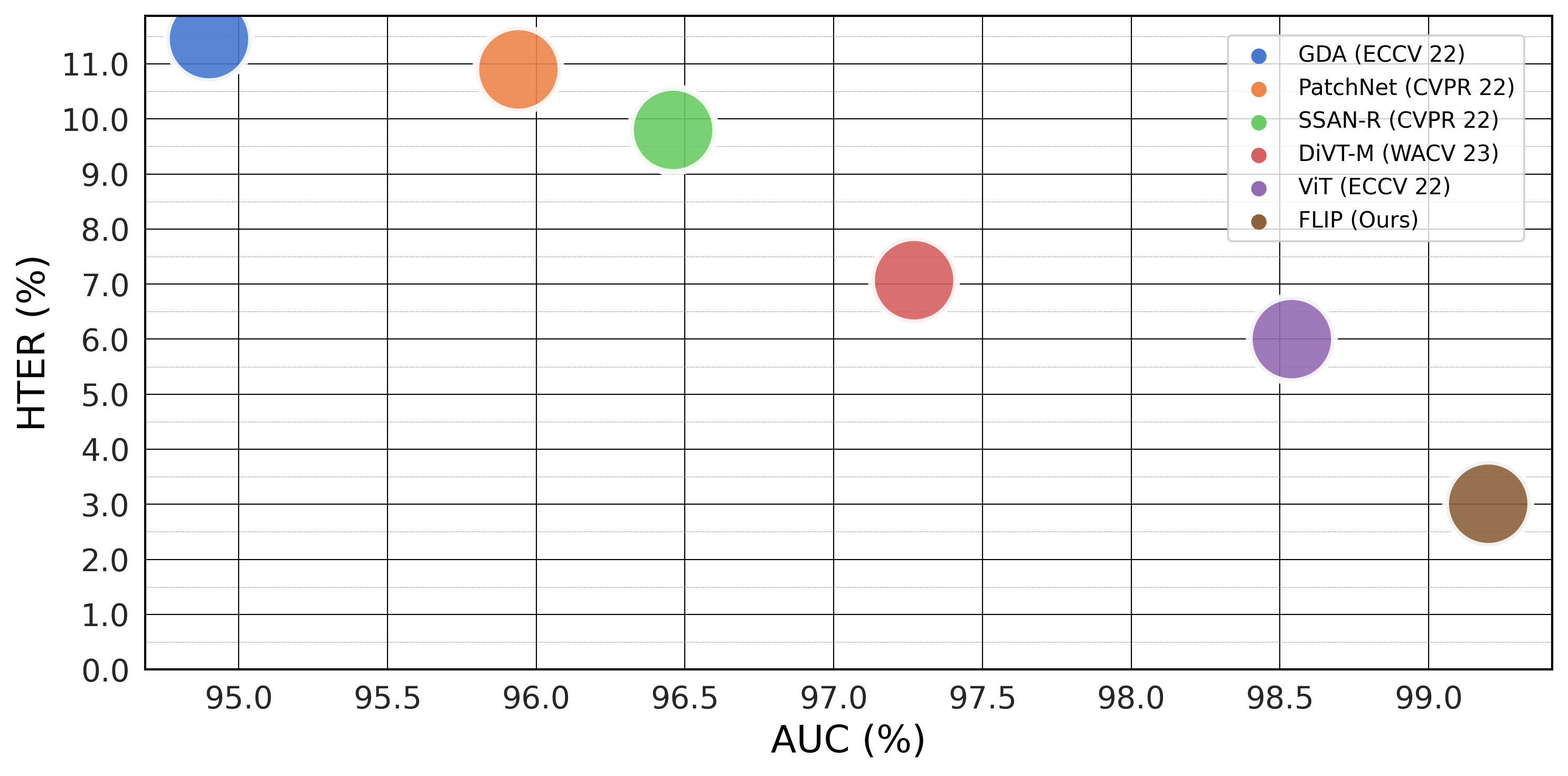}
    \footnotesize
    \caption{Area Under ROC Curve (AUC \%) and Half Total Error Rate (HTER \%) comparison between our proposed method and state-of-the-art (SOTA). Our method achieves the highest AUC ($\uparrow$) performance with the lowest HTER ($\downarrow$) for cross-domain face anti-spoofing on MCIO datasets, surpassing all the SOTA methods.}
    \label{fig:hter_vs_auc}
\end{figure}

The problem of cross-domain FAS has been formulated in different ways in the literature. Unsupervised domain adaptation (UDA) methods \cite{tzeng2017adversarial, ghifary2016deep, hu2018duplex, li2018unsupervised, wang2020unsupervised, wang2020cross, wang2019improving, jia2021unified, zhou2022generative, yue2022cyclically} make use of the unlabeled target domain data and labeled source domain data to learn a generalized decision boundary. Few-shot learning methods \cite{liu2019deep, qin2020learning, perez2020learning, huang2022adaptive} use a small subset of labeled target domain data during training to learn features that adapt well to the target domain. However, both these methods assume access to the target domain either in the form of a large set of unlabeled samples or a few labeled samples, which may not always be available. Domain generalization (DG) methods \cite{shao2019multi, shao2020regularized, chen2021generalizable, liu2021dual, wang2021self, liu2021adaptive, liu2022feature, jia2020single, wang2022domain, 10.1007/978-3-031-20065-6_24, liao2023domain} propose to learn domain-agnostic discriminative features from multiple source domains that generalize to an unseen target domain. While zero-shot learning and DG settings are more challenging, they are more applicable in practice.

Recent works \cite{george2021effectiveness, huang2022adaptive, liao2023domain} have established the effectiveness of vision transformers (ViT) for cross-domain FAS. Since ViTs \cite{dosovitskiy2020image} split the image into fixed-size patches and have the ability to capture long-range dependencies among these patches, they can independently detect the local spoof patterns and aggregate them globally to make an informed decision. However, these methods have two limitations. Firstly, these ViTs are learned using only image data and their learning is guided only by the corresponding image labels, which might not be representative enough. This limits their generalization ability, especially when presented with limited training data. Secondly, they typically require adaptive modules, additional domain labels, or attack-type information to finetune pre-trained weights. This requires explicit network modifications or custom curation of additional information such as attack type or domain labels.

While multimodal vision-language pre-trained (VLP) models have achieved striking zero-shot performance and good generalization in some applications \cite{10.1007/978-3-031-19833-5_29, zhou2022conditional, gu2021open, rasheed2022bridging, zhou2022detecting, li2022language, rao2022denseclip}, there is still a debate on whether incorporating language supervision yields vision models with more generalizable representations \cite{devillers-etal-2021-language, santurkarcaption}. Therefore, the objective of this work is to examine the following questions: (i) Can initialization of ViTs using multimodal pre-trained weights lead to better cross-domain FAS performance compared to ViTs pre-trained only on images?; (ii) Besides leveraging the image encoder of a VLP model, can the text encoder also be utilized to improve the FAS generalization performance?; and (iii) Can the large domain gap and limited training data availability in FAS be surmounted by exploiting  self-supervision techniques during the adaptation of VLP models for the FAS task? The main 
 contributions of this work are as follows:

\begin{itemize}
    \item We show that direct finetuning of a multimodal pre-trained ViT (e.g., CLIP image encoder) achieves better FAS generalizability without any bells and whistles.
    
    \item We propose a new approach for robust cross-domain FAS by grounding the visual representation using natural language semantics. This is realized by aligning the image representation with an ensemble of text prompts (describing the class) during finetuning.
    
    \item We propose a multimodal contrastive learning strategy, which enforces the model to learn more generalized features that bridge the FAS domain gap even with limited training data. This strategy leverages view-based image self-supervision and view-based cross-modal image-text similarity as additional constraints during the learning process.
\end{itemize}

\section{Related Work}
\label{sec:related_works}

\noindent \textbf{Domain Adaptation and Few-shot Learning}: Several methods have been proposed to leverage unlabeled data from the target domain along with labeled source data. One approach is to align the source and target feature distributions either by reducing the Maximum Mean Discrepancy \cite{li2018unsupervised} or by using adversarial domain adaptation \cite{wang2019improving}. Other methods use semi-supervised learning \cite{jia2021unified} and progressive transfer learning strategies \cite{quan2021progressive} to exploit the availability of a few labeled samples from the target domain. In \cite{li2020face}, a FAS model trained with sufficient labeled training data is distilled to application-specific domains for which training samples are scarce. In \cite{zhou2022generative}, cross-domain FAS is treated as a style transfer problem, where target data is transformed to the source domain style via image translation. Vision transformers with ensemble adapter modules and feature-wise transformation layers are employed in \cite{huang2022adaptive} for adapting to the target domain. Pseudo-labeled samples containing domain-invariant liveness features from the source domain and content features from the target domain are generated in \cite{yue2022cyclically} and both these features are disentangled through domain adversarial training. However, all the above methods assume access to the unlabeled/labeled target domain data, which may not always be available.\\

\noindent {\textbf{Domain Generalization}}: The idea of learning a shared generalized feature space for FAS was first proposed in \cite{shao2019multi}, where a multi-adversarial discriminative domain generalization framework was presented. A fine-grained meta-learning-based approach was proposed in \cite{shao2020regularized} by simulating the domain shift during training. The concept of 
separating the features into style and content components to create a stylized feature space was introduced in \cite{wang2022domain}, upon which a contrastive learning strategy is applied emphasizing on liveness-related style information to learn a generalized representation. Recently, vision transformers with two additional losses were used in \cite{liao2023domain}, where one loss enforces the real data from multiple domains to be compact and the other enforces a domain-invariant attack type separation. Though these methods demonstrate promising cross-domain performance, they still require additional information such as attack types and domain labels, or make use of non-trivial auxiliary supervision.\\

\begin{figure*}
    \centering
    \includegraphics[width=0.9\textwidth]{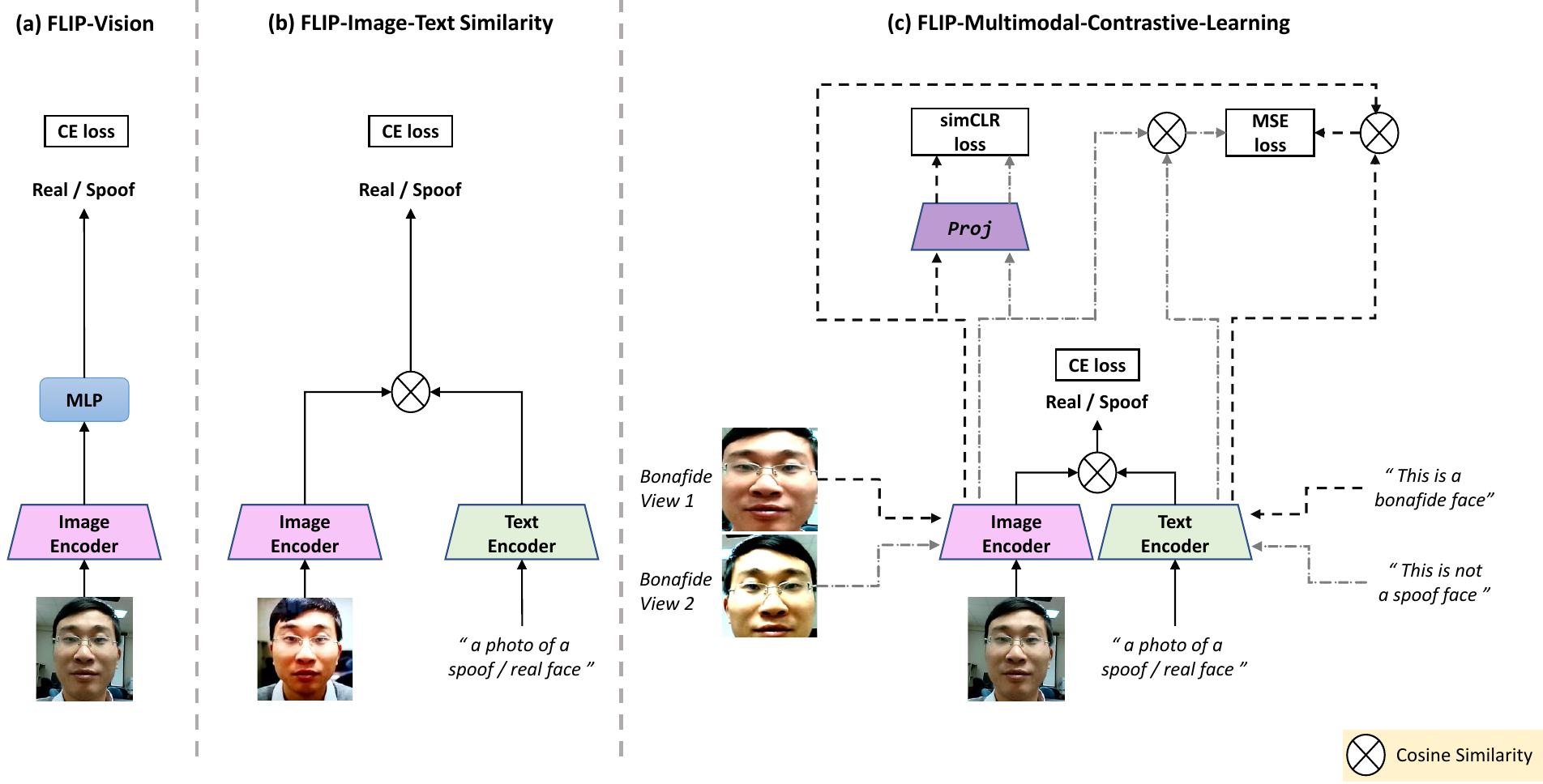}
    \caption{Overview of the proposed FLIP framework for cross-domain face anti-spoofing.}
    \label{fig:flip_framework}
\end{figure*}

\noindent {\textbf{Vision Language Pre-training}}: Vision-language pre-trained (VLP) models encode rich multimodal representations and have demonstrated excellent generalization performance on various downstream applications \cite{10.1007/978-3-031-19833-5_29, zhou2022conditional, gu2021open, rasheed2022bridging, zhou2022detecting, li2022language, rao2022denseclip}. Riding on the success of transformer models \cite{vaswani2017attention, dosovitskiy2020image}, contrastive representation learning \cite{chen2020simple, he2020momentum}, and  web-scale training datasets \cite{jia2021scaling, radford2021learning}, several VLP models have been proposed recently to learn joint image-text representations \cite{radford2021learning, jia2021scaling, zhai2022lit, yao2021filip, yuan2021florence}. However, the issue of whether language supervision enhances the generalizability of vision models is still being debated \cite{devillers-etal-2021-language, santurkarcaption}. In this work, we use contrastive language-image pre-training (CLIP) \cite{radford2021learning} as the base VLP model.\\

\section{Proposed Method}
\label{sec:method}

\noindent The goal of cross-domain FAS is to achieve high presentation attack detection accuracy on out-of-distribution face datasets containing bonafide images and presentation attacks. In the many-to-one DG setting, the model is learned from a set of $N$ different source domain datasets $\mathcal{S} = \{\mathcal{S}_1, \mathcal{S}_2, \cdots, \mathcal{S}_N\}$ and evaluated on a single target domain dataset $\mathcal{T}$.  In the one-to-one DG setting, the model is trained on images from a single source domain $\mathcal{S}_i$ to generalize to the target domain. Let $I_{\mathcal{D}}^{r}$ denote a real (bonafide) face image from domain $\mathcal{D} \in (\mathcal{S} \cup \mathcal{T})$. Similarly, let $I_{\mathcal{D}}^{s}$ represent a spoof (presentation attack) image from $\mathcal{D}$.

We propose a framework called \textbf{F}ace Anti-Spoofing with \textbf{L}anguage-\textbf{I}mage \textbf{P}retraining (FLIP) for cross-domain FAS (see Figure \ref{fig:flip_framework}). The proposed framework uses CLIP \cite{radford2021learning} as the base model and is finetuned using different strategies to obtain three variants: FLIP-Vision (FLIP-V), FLIP-Image-Text Similarity (FLIP-IT), and FLIP-Multimodal-Contrastive-Learning (FLIP-MCL). We first outline the working of the base model before describing the variants.

\subsection{Contrastive Language-Image Pre-Training}
\label{subsec:CLIP}
CLIP \cite{radford2021learning} is trained using millions of image-text pairs sourced from the internet. CLIP encodes the input image $I \in \mathbb{R}^{H \times W \times 3}$ and the corresponding text description $t$ into a shared embedding space as detailed below.

\noindent \textbf{Image Encoder}: The image encoder is a vision transformer $\mathcal{V}$ consisting of $K$ transformer blocks $\{\mathcal{V}_k\}_{k=1}^{K}$. To encode the input image $I$, it is first split into $M$ fixed-size patches and these patches are projected linearly into patch embeddings $\vect{e}_0 \in \mathbb{R}^{M \times d_v}$. Patch embeddings $\vect{e}_{k-1}$ are then input to the $k^{\text{th}}$ transformer block $(\mathcal{V}_{k})$ after appending a learnable class token $\text{c}_{k-1}$, and processed through the $K$ transformer blocks sequentially.
\vspace{-1.8em}

\begin{align*}
    [{\text{c}}_k, \vect{e}_k] &= \mathcal{V}_{k}([\text{c}_{k-1}, \vect{e}_{k-1}])  \qquad k= 1, 2, \cdots, K.
\end{align*}

\noindent The final image representation $\vect{x}$ is obtained by linearly projecting the class token $\text{c}_{K}$ from the last transformer block $(\mathcal{V}_{K})$ into a shared vision-language space via \texttt{ImageProj}:
\vspace{-1.8em}

\begin{align*}
    \vect{x} &= \texttt{ImageProj}({\text{c}}_{K}) \ \ \ \qquad \vect{x} \in \mathbb{R}^{d_{vl}}.
\end{align*}

\noindent \textbf{Text Encoder}: The text encoder $\mathcal{L}$ generates feature representations for the description $t$ by first tokenizing the words and then projecting them into word embeddings 
$\vect{w}_0 =[w_{0}^{1}, w_{0}^{2}, \cdots, w_{0}^{Q}] \in \mathbb{R}^{Q \times d_{l}}$.
At each stage, $\vect{w}_{k-1}$ is input to the $k^{\text{th}}$ transformer block $(\mathcal{L}_{k})$ to obtain
\begin{align*}
    \vect{w}_k = \mathcal{L}_{k}(\vect{w}_{k-1}) \ \ \ \qquad k= 1, 2, \cdots, K.
\end{align*}
The final text representation $\vect{z}$ is obtained by projecting the text embeddings corresponding to the last token of the last transformer block $(\mathcal{L}_K)$ into a shared vision-language latent space via \texttt{TextProj}.
\begin{align*}
    \vect{z} = \texttt{TextProj}({w_{K}^{Q}}) \ \ \ \ \ \qquad \vect{z} \in \mathbb{R}^{d_{vl}}.
\end{align*}

\noindent The CLIP model has been pre-trained using a contrastive loss that maximizes the cosine similarity of the image ($\vect{x}$) and text ($\vect{z}$) embeddings of $n$ corresponding (image, text) pairs in a batch while minimizing the cosine similarity of the embeddings of the ($n^2-n$) incorrect pairings.

\subsection{FLIP-Vision}
Representations produced by CLIP have shown impressive out-of-the-box performance for many downstream vision applications based on natural images such as classification \cite{10.1007/978-3-031-19833-5_29, zhou2022conditional}, object detection \cite{gu2021open, rasheed2022bridging, zhou2022detecting}, and segmentation \cite{li2022language, rao2022denseclip}. However, these features cannot be directly used for the FAS task, which requires identifying subtle variations among similar face images. Hence, we first finetune only the vision backbone for FAS and refer to this approach as FLIP-Vision (FLIP-V). In this method, we take a pre-trained CLIP model and use only its image encoder $\mathcal{V}$ and discard the text encoder $\mathcal{L}$. This gives us a simple ViT initialized with language-image pre-trained weights. Given a batch of balanced images from $N$ source domains, we use the image encoder to extract the class token $(\text{c}_{K})$ from the last transformer block $(\mathcal{V}_{K})$ prior to \texttt{ImageProj}. This class token is then passed to a multi-layer perceptron (MLP) classification head, to decide if the input image is spoof or real. The image encoder and the MLP head are updated using the standard cross entropy loss $L_{ce}$.

\begin{table}[htbp]
    \centering
    \scalebox{0.6}{
        \begin{tabular}{|c|c|c|}
        \hline
         \rowcolor{mygray} \textbf{Prompt No.} & \textbf{Real Prompts} & \textbf{Spoof Prompts } \\
         \hline
         P1 & This is an example of a real face & This is an example of a spoof face \\
         P2 & This is a bonafide face &  This is an example of an attack face\\
         P3 & This is a real face & This is not a real face \\
         P4 & This is how a real face looks like & This is how a spoof face looks like \\
         P5 & A photo of a real face & A photo of a spoof face \\
         P6 & This is not a spoof face & A printout shown to be a spoof face\\
         \hline
    \end{tabular}
    }
    \vspace{0.5em}
    \caption{Natural language descriptions (context prompts) of the real and spoof classes used to guide the FLIP-IT model.}
    \label{tab:ensemble_texts}
\end{table}

\input{tables/table_mcio}


\subsection{FLIP-Image-Text Similarity}
\label{subsubsec:flip_it}
In \textbf{FLIP}-\textbf{I}mage-\textbf{T}ext similarity, we obtain the prediction with the help of language supervision instead of using the MLP head. Specifically, we leverage textual prompts/descriptions corresponding to the real and spoof classes (denoted as $t_r$ and $t_s$, respectively), whose feature representations are computed using the text encoder $\mathcal{L}$. The cosine similarity between the image representation ($\vect{x}$) and text representations corresponding to the two classes ($\vect{z}_r$ and $\vect{z}_s$) is computed, resulting in two values for every image in the batch. These similarity values are considered as class logits and passed to the cross entropy loss computation.

During inference, the predicted class $\hat{y}$ is determined by the class description having the highest cosine similarity score ($sim(\cdot,\cdot)$) with the given image $I$. Hence,
\begin{align*}
\label{eq:flip_it_test}
p(\hat{y}|x) = \frac{\text{exp}(sim(\vect{x}, \vect{z}_{\hat{y}})/\tau)}{\text{exp}(sim(\vect{x}, \vect{z}_{r})/\tau)+\text{exp}(sim(\vect{x}, \vect{z}_{s})/\tau)},
\end{align*}
\noindent where $\tau$ is the temperature parameter and $\hat{y} \in \{r,s\}$ is the predicted class label. To account for the limited availability of training data, we align each image to an ensemble of class descriptions/ prompts called \emph{context prompts}. We consider $P$ descriptions per class and compute the text representation $\vect{z}$ for each description. An average of these representations ($\vect{\bar{z}}$) gives an ensemble of the context in the embedding space. Aligning the image with a multitude of natural language class descriptions enables the model to learn class-specific clues. The specific language descriptions used to describe the real and spoof classes are provided in Table \ref{tab:ensemble_texts}.

\subsection{FLIP-Multimodal-Contrastive-Learning}
\noindent  In \textbf{FLIP}-\textbf{M}ultimodal-\textbf{C}ontrastive-\textbf{L}earning (FLIP-MCL), we propose an additional multimodal contrastive learning objective to further enhance the generalizability of the extracted features and surmount the domain-gap and limited-data problems. This approach is motivated by the tremendous promise of contrastive view-based self-supervised learning methods \cite{chen2020simple, zbontar2021barlow, bardes2022vicreg}. In addition to the cross-entropy loss applied on the cosine similarity logits as described in Section \ref{subsubsec:flip_it}, we also apply self-supervised \texttt{simCLR} loss and mean squared error (\texttt{MSE}) loss. While the \texttt{simCLR} loss is applied on a pair of image views, the \texttt{MSE} loss enforces consistency between pairs of image-text views. 

For the \texttt{simCLR} loss, we follow the approach in \cite{chen2020simple} to create two views (denoted as $I^{v_1}$ and $I^{v_2}$) of the given image $I$ by applying different transformations. The features corresponding to the two transformed images are extracted using the image encoder $\mathcal{V}$ and further projected using a non-linear projection network $\mathcal{H}$. Finally, a contrastive loss is applied on the projected features. 
\begin{align*}
    \vect{x}^{v_1} = \mathcal{V}(I^{v_1}) , \ \ \ \ \vect{x}^{v_2} = \mathcal{V}(I^{v_2}) 
\end{align*}
\vspace{-2em}
\begin{align*}
    \vect{h}_1 = \mathcal{H}(\vect{x}^{v_1}) \ \ , \ \ h_2 = \mathcal{H}(\vect{x}^{v_2}) \ \ \ \quad \vect{h}_1, \vect{h}_2 \in \mathbb{R}^{d_{h}}.
\end{align*}
\vspace{-2em}
\begin{align*}
    L_{simCLR} = \texttt{simCLR}(\vect{h}_1, \vect{h}_2)
\end{align*}
For the \texttt{MSE} loss, we first randomly sample two different prompts from the ground-truth class and get their text representations $\vect{z}^{v_1}$ and $\vect{z}^{v_2}$. We now have two image views and two text views. For each pair of image-text views, we compute the cosine similarity score between the image and text representations and enforce the consistency between the two similarity scores.
\begin{align*}
    L_{mse} =  (sim(\vect{x}^{v_1}, \vect{z}^{v_1}) - sim(\vect{x}^{v_2}, \vect{z}^{v_2}))^2
\end{align*}

\noindent We define the joint training objective as:
\begin{align*}
    L_{mcl} &= L_{ce} + L_{simCLR} + L_{mse}
\end{align*}

\noindent We follow the same cosine similarity method described in Section \ref{subsubsec:flip_it} for inference.

\input{tables/table_wcs}
\input{tables/table_one_to_one_mcio}

\section{Experiments}
\label{sec:experiments}
\subsection{Experimental Setup}
\noindent \textbf{Datasets and DG Protocols}:
We evaluate our method on three different protocols. Following \cite{huang2022adaptive}, we set up the first two protocols as a leave-one-domain-out testing protocol, where each dataset is considered as a domain and we evaluate the cross-domain performance on the left-out domain. In \textbf{Protocol 1}, we evaluate on the widely used cross-domain FAS benchmark datasets, MSU-MFSD \textbf{(M)} \cite{7031384}, CASIA-MFSD \textbf{(C)} \cite{6199754}, Idiap Replay Attack \textbf{(I)} \cite{6313548}, and OULU-NPU \textbf{(O)} \cite{7961798}. For example, \textbf{OCI} $\rightarrow$ \textbf{M} represents the scenario where \textbf{O}, \textbf{C}, and \textbf{I} datasets are considered as source domains and \textbf{M} is the target domain. In \textbf{Protocol 2}, we evaluate our method on the large-scale FAS datasets, WMCA \textbf{(W)} \cite{8714076}, CASIA-CeFA \textbf{(C)} \cite{9423056, liu2021cross}, and CASIA-SURF \textbf{(S)} \cite{8995504, zhang2019dataset}. To further evaluate the performance in the low-data regime, we follow \cite{yue2022cyclically} and set up \textbf{Protocol 3} as a single-source-to-single-target protocol. We use the \textbf{M}, \textbf{C}, \textbf{I}, and \textbf{O} datasets, where each source domain will have 3 combinations, one each with the other domains, giving us a total of 12 different scenarios. In each of the three protocols, similar to \cite{huang2022adaptive}, we include CelebA-Spoof \cite{zhang2020celeba} as the supplementary training data to increase the diversity of training samples. \\

\noindent \textbf{Implementation Details}:
We crop and resize the face images to $224 \times 224 \times 3 $ and split them into a patch size of $16 \times 16$. For the image encoder, we use the ViT variant of the CLIP model. For the text input, we have curated a set of custom text prompts for each of the real and spoof classes as shown in Table \ref{tab:ensemble_texts}. We use the Adam optimizer and set the initial learning rate to $10^{-6}$ and weight decay to $10^{-6}$. For each domain, we set a batch size of 3 in \textbf{Protocol 1} and \textbf{Protocol 3} and a batch size of 8 in \textbf{Protocol 2}. For FLIP-V we use a two-layer MLP head containing fully-connected layers of dimensions 512 and 2 respectively. The dimensionality of the image representation is $d_v = 768$ and the dimension of the shared vision-language embedding space is $d_{vl} = 512$. For all the 3 variants of our approach, we train for 4000 iterations. In FLIP-V we update all the layers of the image encoder and MLP, for FLIP-IT we update all the layers of the image and text encoders, and for FLIP-MCL we update all the layers of the image encoder, text encoder, and the non-linear projection network $\mathcal{H}$. In FLIP-MCL, $\mathcal{H}$ consists of 3 linear layers of dimensions 512, 4096, and 256, and the first two layers are followed by BatchNorm and ReLU.\\

\noindent \textbf{Evaluation Metrics}:
Following \cite{huang2022adaptive},  we evaluate the model performance using the Half Total Error Rate (HTER), Area Under the Receiver Operating Characteristic Curve (AUC), and True Positive Rate (TPR) at a fixed False Positive Rate (FPR). Unlike most prior works that simply report the best result over a single trial, we run each of our experiments 5 times with different random seeds and report the mean HTER, AUC, and TPR@FPR=1\% in all the results. The standard deviation of the performance metrics is reported in the supplementary material along with the statistical hypothesis testing results. \\

\noindent \textbf{Baseline Methods}:
The closest and state-of-the-art (SOTA) baseline methods for the proposed FLIP framework are ViT-based FAS methods reported in \cite{huang2022adaptive} and \cite{liao2023domain}. While \cite{huang2022adaptive} reports both zero-shot and five-shot performance, it uses only vanilla ViT for the zero-shot case, but both vanilla and adaptive ViTs (ViTAF) for the five-shot case. Only zero-shot performance is considered in \cite{liao2023domain}. Note that zero-shot refers to the setting where no sample from the target domain is used during training, while five-shot refers to the setting where 5 labeled samples from the target domain are used during training. 

\subsection{Cross-domain FAS Performance}

\noindent Table \ref{tab:cross_db_mcio}, Table \ref{tab:cross_db_wcs}, and Table \ref{tab:result_da} report the zero-shot cross-domain performance for \textbf{Protocol 1}, \textbf{Protocol 2}, and \textbf{Protocol 3}, respectively. We can further extend the proposed FLIP framework for the five-shot setting following techniques similar to \cite{huang2022adaptive}, and the corresponding five-shot results are provided in the supplementary material.

\noindent \textbf{Comparison of proposed training strategies}: Firstly, we analyze the performance of the FLIP-V variant, which is obtained by simple finetuning of a multimodal pre-trained ViT. The results in Tables \ref{tab:cross_db_mcio}, \ref{tab:cross_db_wcs}, and \ref{tab:result_da} show that even this simple strategy can achieve SOTA performance (in terms of average HTER) on all three protocols, demonstrating the zero-shot transfer capabilities of VLP models. Note that this result belies claims in \cite{huang2022adaptive} and \cite{george2021effectiveness} that full finetuning of a pre-trained ViT image encoder inhibits its generalizability. In two of the three protocols considered (Protocols 1 and 3), the FLIP-IT variant outperforms the FLIP-V variant. This illustrates the power of natural language supervision in generating more generalizable representations, especially when the training data is limited. Even in the case of Protocol 2, the FLIP-IT variant generalizes better than FLIP-V in two of the three scenarios (see Table \ref{tab:cross_db_wcs}), with poor performance only in the \textbf{CW} $\rightarrow$ \textbf{S} case. Finally, the proposed FLIP-MCL variant significantly outperforms all the SOTA methods for all three protocols in the zero-shot setting. In the case of Protocol 1, the zero-shot performance of FLIP-MCL is better than even the five-shot performance of the SOTA ViTAF. This clearly demonstrates the effectiveness of the proposed multimodal contrastive learning strategy.

\noindent \textbf{Cross-domain performance in Protocol 1}: The FLIP framework outperforms SOTA zero-shot methods in three out of four target domains (C=+5.2, I=+0.76, O=+5.16) and five-shot methods in two out of four target domains (C=+0.86, O=+3.08) by large margins. We observe that the performance drop in M (-3.37) is primarily due to the real samples being categorized as presentation attacks, thereby increasing the false negative error rate. Compared to zero-shot methods, we can also observe huge gains in TPR@FPR=1\% in three out of the four domains (C=+11.43, I=+33.08, O=+22.98). \\

\noindent \textbf{Cross-domain performance in Protocol 2}: The proposed FLIP framework performs better than zero-shot ViT in all three domains (W=+3.52, C=+1.47, and S=+1.64) in terms of HTER. In terms of TPR@FPR=1\%, we are able to see high gains of +10.25, +11.41, and +8.01 for the target domains W, C, and S respectively. Compared to Protocol 1, Protocol 2 has much more subjects ($>$ 1000 in CASIA-CeFA/SURF, compared to $\approx$50 in MCIO) and richer environmental variations, which once again proves the effectiveness of our approach in learning generalized features across different data regimes. \\

\noindent \textbf{Cross-domain performance in Protocol 3}: In the challenging single-source to single-target setting, our framework outperforms (in terms of average HTER) SOTA methods by a large margin of +8.36. Specifically, for the target domain O, we observe huge HTER improvements of +26.0, +25.7, and +21.65, when taking C, I, and O as the source domains respectively.  Also, for the target domain C, we observe huge improvements of +11.22, +8.61, and +18.91, when taking I, M, and O as the source domains. For the target domain M, we observe improvements of +0.95, and +2.38, for source domains C and I, except for O (-2.1). For the target domain I, we observe that \cite{yue2022cyclically} does better for the source domains C and M, but for source domain O, our framework is able to perform on par. These results demonstrate that the FLIP-MCL method can learn strong generalizable features that could handle adverse limited-data and domain-gap problems. 

\subsection{Ablation Studies}

\noindent \textbf{Comparing various ViT initialization methods for FAS:}
To extend our observation regarding the effect of initialization on FAS generalizability, we take ViT pre-trained with different methods and show the comparative performance in Table \ref{tab:ablation_initialization}. Specifically, we adopt the ViT training strategy proposed in \cite{huang2022adaptive} and a) train from scratch without any pre-trained weights, b) initialize with self-supervised BeIT \cite{bao2022beit} pre-training weights, c) initialize with ImageNet pre-trained weights \cite{huang2022adaptive} and d) initialize with multimodal CLIP \cite{radford2021learning} pre-trained weights. It can be seen that multimodal pre-trained initialization achieves better FAS generalizability compared to other initialization methods due to their ability to encode rich multimodal representations, serving as a base for all the experiments aligning image and text representations. \\
\input{tables/table_ablation_pretraining_compare}
\begin{figure*}
    \centering
    \includegraphics[width=1\textwidth]{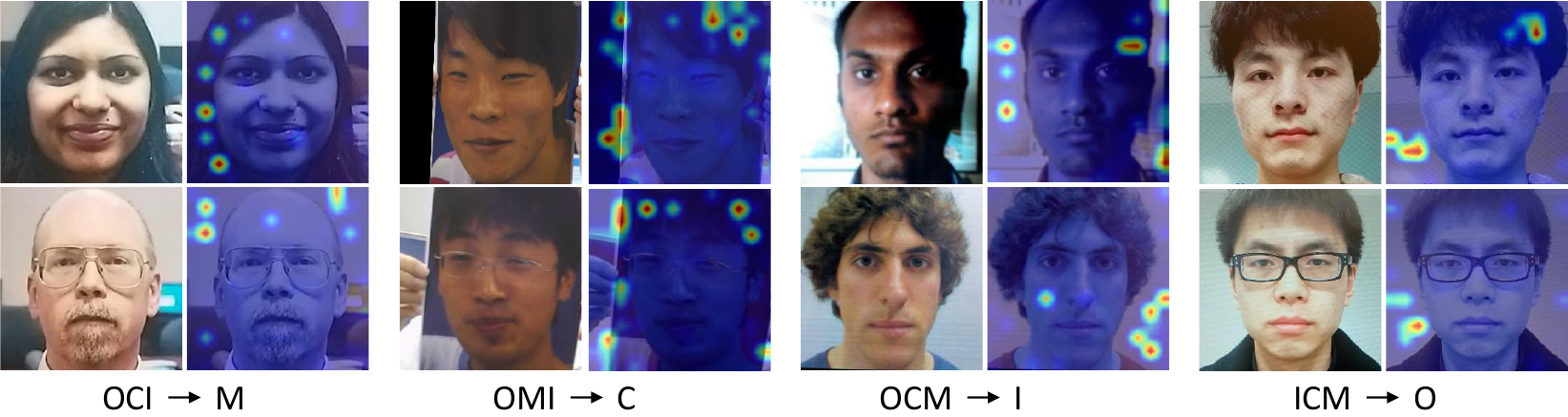}
    \footnotesize
    \caption{\textbf{Attention maps on spoof images from different scenarios in Protocol 1:}  We observe that the attention highlights are on the spoof-specific clues such as paper texture (M), edges of the paper (C), and moire patterns (I and O).}
    \label{fig:attention_maps}
    \vspace{0.5em}
\end{figure*}
\begin{figure*}
    \centering
    \includegraphics[width=1\textwidth]{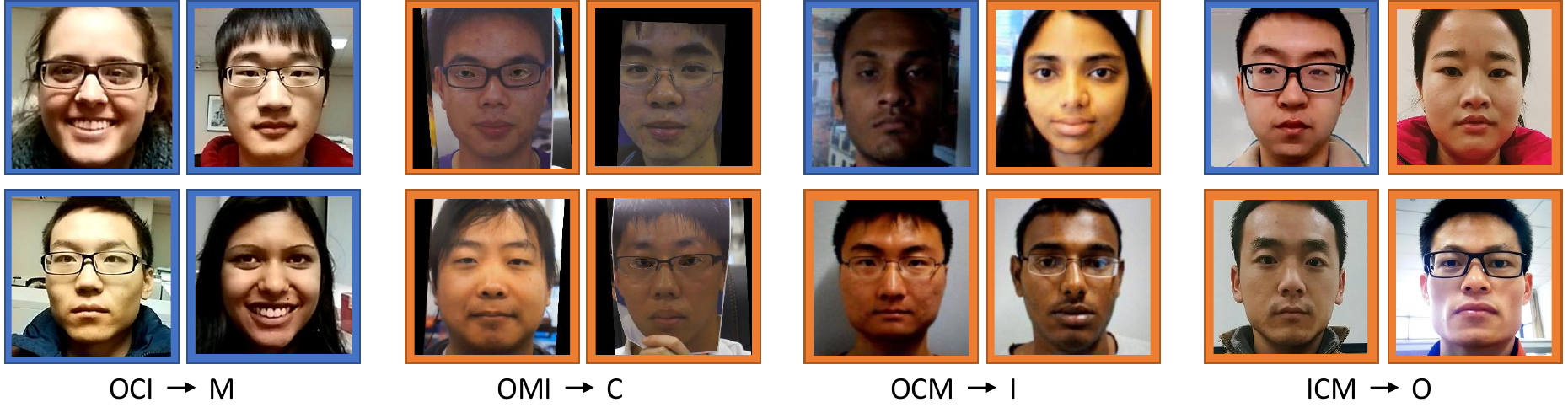}
    \footnotesize
    \caption{\textbf{Mis-Classified Examples in Protocol 1:} Blue boxes indicate real faces mis-classified as spoof. Orange boxes indicate spoof faces mis-classified as real.}
    \label{fig:misclassified_analysis}
\end{figure*}

\noindent \textbf{Impact of different text prompts:}
In Table \ref{tab:ablation_prompts}, we compare the effect of different text prompts in guiding the classification decision. It can be seen that different text prompts perform well for different cross-domain scenarios and it is difficult to choose a single prompt that works well across all the cases. Creating a list of different prompts for real and spoof classes is relatively easier and the performance of ensemble prompts shows that it is able to capture the best representation from each prompt while eliminating any inherent noise. This validates our idea of aligning the image representation to an ensemble of class prompts to learn generalized representations.\\
\input{tables/table_ablation_effect_of_prompts}
\input{tables/supplementary/suppl_param_sen_analysis}
\noindent\textbf{Contribution of different loss terms:}
We weight the different components of the joint training loss of FLIP-MCL as follows: $L_{mcl} = \alpha L_{ce} + \beta L_{simCLR} + \gamma L_{mse}$. A sensitivity analysis based on the tuple $(\alpha,\beta,\gamma)$ is provided in Table \ref{tab:suppl_param_sens_analyis}.  Note that self-supervised losses $L_{simCLR}$ and $L_{mse}$ provide regularization in combination with the supervised cross-entropy loss $L_{ce}$. As we increase the importance of $L_{simCLR}$ and $L_{mse}$ losses (e.g., $(1,2,2)$ and $(1,5,5)$), it reduces the overall performance. This is expected because these settings decrease the contribution of $L_{ce}$ during training. Similarly, the performance degrades when $\beta=0$ or $\gamma=0$, verifying that the self-supervised losses indeed facilitate better generalization.\\

\begin{figure*}
    \centering
    \includegraphics[width=1\textwidth]{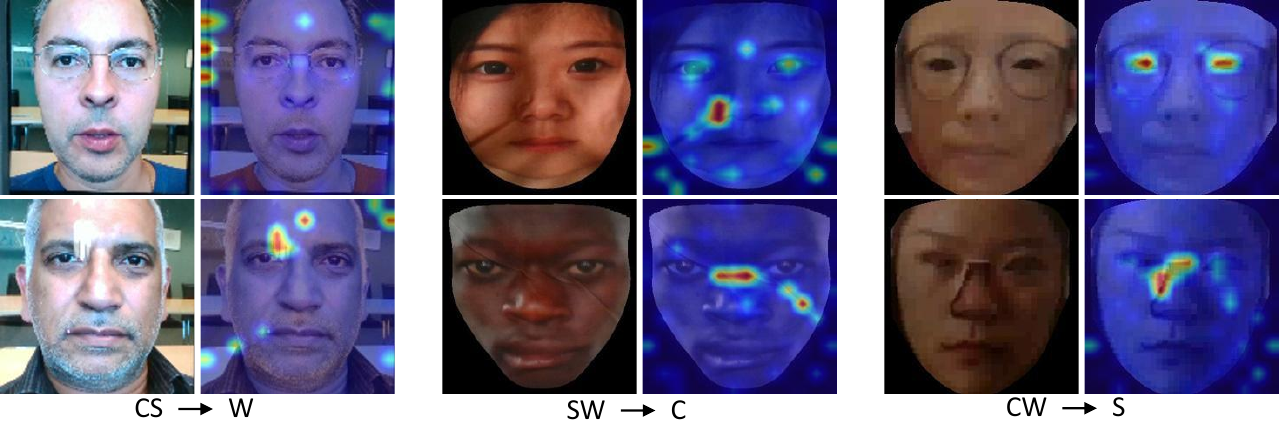}
    \footnotesize
    \caption{\textbf{Attention maps on spoof images from different scenarios in Protocol 2:} We observe that the attention highlights are on the spoof-specific clues such as screen edges/ screen reflection (W), wrinkles in printed cloth (C), and cut-out eyes/nose (S).}
    \label{fig:suppl_wcs_attention_maps}
    \vspace{0.5em}
\end{figure*}
\begin{figure*}
    \centering
    \includegraphics[width=1\textwidth]{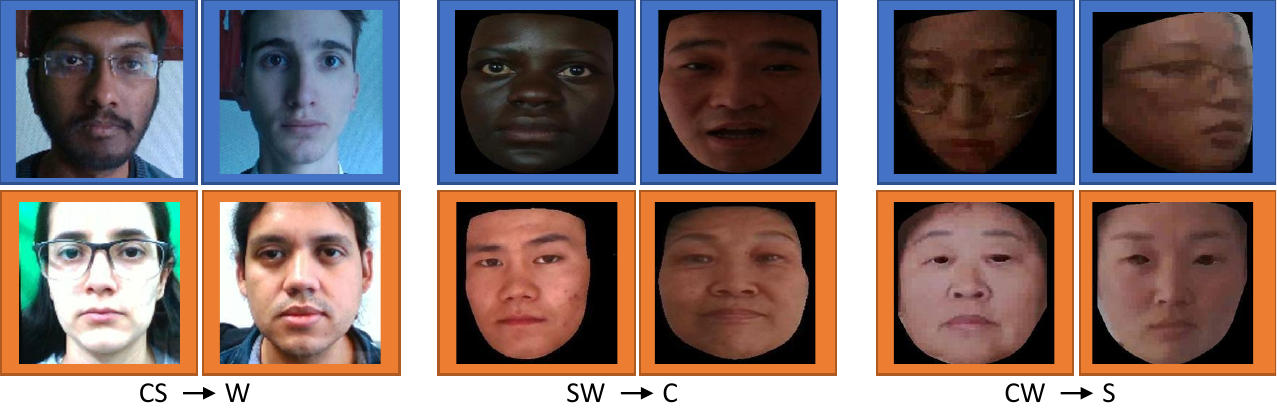}
    \footnotesize
    \caption{\textbf{Mis-Classified Examples in Protocol 2}: Blue boxes indicate real faces mis-classified as spoof. Orange boxes indicate spoof faces mis-classified as real.}
    \label{fig:suppl_wcs_mis_classify}
\end{figure*}

\subsection{Visualization}
\noindent \textbf{Attention maps:}
In Figure \ref{fig:attention_maps} and Figure \ref{fig:suppl_wcs_attention_maps}, we use \cite{Chefer_2021_ICCV} to show the visual attention maps of the FLIP-MCL model on the spoof samples in \textbf{Protocol 1} and \textbf{Protocol 2} respectively. We can observe that our model is able to effectively localize the spoof patterns in each of the spoof domains to make the classification decision. In \textbf{Protocol 1} the datasets contain only print and replay attacks. We observe from the figure that the attention highlights are on the spoof-specific clues such as paper texture (M), edges of the paper (C), and moire patterns (I and O). In \textbf{Protocol 2}, for the CS $\rightarrow$ W scenario, we observe that the model focuses on spoof clues such as the edges of the paper/screen or the reflection on the screen. For the SW $\rightarrow$ C scenario, we observe that the model focuses on the region with cloth wrinkles. For the CW $\rightarrow$ S scenario, we observe that the model focuses on the cut region of the nose, or eyes.\\

\noindent \textbf{Mis-Classified examples:}
In Figure \ref{fig:misclassified_analysis}, we show examples of images being mis-classified in \textbf{Protocol 1}. It is interesting to observe that for the OCI $\rightarrow$ M scenario, there are no false positive cases. i.e., none of the spoof samples have been predicted as real. However, as shown in Figure \ref{fig:misclassified_analysis}, some of the bonafide samples are mis-classified as spoof due to low image resolution and lighting variations, causing the performance to drop as shown in Table \ref{tab:cross_db_mcio}. In contrast, for the OMI $\rightarrow$ C scenario, we observe that none of the real samples are mis-classified as spoof, but a few high-resolution spoof samples are mis-classified as real. This could be due to the presence of high-resolution images from OULU (O) in training. For the OCM $\rightarrow$ I scenario, we observe that only 0.62\% of the real samples are incorrectly classified. For the spoof samples, the mis-classification could be attributed to the adverse change in lighting conditions. For the ICM $\rightarrow$ O scenario, we again observe that a very low percentage (0.2\%) of the real samples are mis-classified as spoof. Samples in O have higher resolution compared to the other datasets as shown, and this could be attributed to mis-classifying spoof as real.

In Figure \ref{fig:suppl_wcs_mis_classify}, we show the examples of images being mis-classified in \textbf{Protocol 2}. For the CS $\rightarrow$ W scenario, we observe that some real samples are mis-classified as spoof due to the texture in the background region, which is identified as a moire spoof pattern visible in replay attacks. For the spoof samples being mis-classified as real, we observe that there are no clear visible spoof clues on these print and replay mediums. For the SW $\rightarrow$ C scenario, we observe that real samples in darker lighting conditions or a few faces with darker skin tones are mis-classified as spoof. The spoof sample mis-classification can be attributed to a realistic cloth print or print attack with no visible spoof clues, making it challenging for the model. For the CW $\rightarrow$ S scenario, we observe that most of the samples are of poor image resolution with a lot of pixelization. The real samples being mis-classified as spoof is either due to a) Pixelization, b) extreme pose changes, or c) darker lighting conditions. Some of the spoof samples that have higher resolution compared to the other samples get mis-classified as real.

\section{Conclusion}
In this work, we have shown that vision transformer models learned using vision-language pre-training (e.g., CLIP) have excellent generalization ability for the face anti-spoofing task, compared to their counterparts trained only on images. The rich multimodal representations learned by these models enable them to work well, even if only the image encoder is finetuned and used for presentation attack detection. On top of this baseline, we have shown that aligning the image representations to text representations produced by the text encoder further boosts generalizability. Using multimodal contrastive learning also enhances the generalizability across data regimes and domain gaps. The limitation of the later approaches is the additional computational overhead involved in invoking the text encoder during training. In the future, we plan to explore if these conclusions hold for other VLP foundation models. Prompt learning is also a potential way to further improve performance.

\newpage
{\small
\bibliographystyle{ieee_fullname}
\bibliography{egbib}
}

\newpage
\appendix

\begin{center}
    \textbf{\Large Supplementary Material}
\end{center}

In this supplementary material, we provide the experimental results for extending the FLIP framework to the 5-shot setting in section \ref{sec:flip_few_shot}. In section \ref{sec:stat_sig}, we provide the results of the statistical significance test. In section \ref{sec:comp_comp}, we compare the computational complexity and network parameters between all the methods in the FLIP framework. 
In section \ref{sec:additional_experiment_unseen_spoof_type}, we present the results of evaluating on unseen spoof type. 
\section{Performance of FLIP in 5-shot setting}
 \label{sec:flip_few_shot}
Following \cite{huang2022adaptive}, we evaluate the FLIP framework under the 5-shot setting, where 5 labeled samples from the target domain are available during training to help bridge the domain gap. Tables \ref{tab:suppl_cross_db_mcio}, \ref{tab:suppl_cross_db_wcs}, and \ref{tab:suppl_result_da} report the cross-domain 0-shot and 5-shot performance of Protocol 1, 2, and 3, respectively. \\

\noindent \textbf{5-shot performance in Protocol 1:} We observe that all the 3 methods from the FLIP framework outperform the baseline 5-shot performance. Notably, for the O protocol (where the target samples have higher image resolution and are 4 times larger than all the source domains combined), we observe a large HTER gain of +3.85\%. This demonstrates that our method is able to effectively adapt to larger unknown domains with very few samples ($\approx$ 0.16\% of target domain samples in O). \\

\noindent \textbf{5-shot performance in Protocol 2:} Similar to \textbf{Protocol 1}, we observe that our framework outperforms the baseline 5-shot methods by a huge margin of +2.43\% in terms of average HTER. Notably, for the C and S protocols (which contain more than 1000 identities and have large illumination variations), we observe HTER gains of +3.87\% and +5.4\% respectively. This demonstrates the effectiveness of our method in adopting to unknown distributions containing diverse samples, with just a few labeled samples (0.08\% for C and 0.1\% for S). \\

\noindent \textbf{5-shot performance in Protocol 3:} To make a fair comparison, we implement the baseline ViTAF* method \cite{huang2022adaptive} and extend it to Protocol 3 under the 5-shot setting. We observe that the performance of the FLIP framework in the 5-shot setting outperforms its 0-shot counterpart. Additionally, the 5-shot FLIP framework also outperforms 5-shot ViTAF* by a margin of +2.26\% (HTER). This corroborates our previous observations on our approach's effectiveness in adapting to unknown domains with a few labeled samples. \\


\input{tables/supplementary/suppl_mcio}

\vspace{1em}
\input{tables/supplementary/suppl_wcs}
\input{tables/supplementary/suppl_one_to_one}
\input{tables/supplementary/suppl_computational_complexity}

\section{Statistical Significance Test}
\label{sec:stat_sig}

Most prior works in cross-domain FAS simply report the best result over a single trial. However, a fair comparison of different methods is possible only when the statistical variations are taken into account. Hence, we run each of our experiments 5 times with different random seeds and report the mean and standard deviation of all the metrics in Tables \ref{tab:suppl_cross_db_mcio}, \ref{tab:suppl_cross_db_wcs}, and \ref{tab:suppl_result_da}. For each of the three protocols, we observe that the standard deviation of the proposed method is low, indicating stable performance across multiple runs.\\

Furthermore, for \textbf{Protocol 1} and \textbf{Protocol 2}, we perform a one-sided pair-wise t-test to evaluate whether the proposed method outperforms the baseline. Specifically, we compare the proposed FLIP-MCL against ViT in the 0-shot setting and against ViTAF* \cite{huang2022adaptive} in the 5-shot setting. The null hypothesis is that there is no statistically significant difference between FLIP-MCL and the baseline, while the alternate hypothesis is that FLIP-MCL is better. In \textbf{Protocol 1}, we find that the null hypothesis is rejected in three out of four scenarios, failing only for M (for both the 0-shot and 5-shot setting). For \textbf{Protocol 2}, the null hypothesis is rejected for all three scenarios in the 0-shot setting. However, for the 5-shot setting, the null hypothesis is rejected for two out of three scenarios failing only in W. These results clearly demonstrate that FLIP-MCL is superior to the baseline methods and the better generalization performance is not due to cherry picking of best trials.

\section{Computational Complexity}
\label{sec:comp_comp}
We present the model size, training, and inference time computational complexity (computed on an NVIDIA Quadro RTX 6000) in Table \ref{tab:suppl_computaional_comp_params}. Kindly note that our image encoder (FLIP-V) is similar to \cite{huang2022adaptive} except that it is pre-trained using CLIP. However, FLIP-IT and FLIP-MCL require an additional text encoder during training. Furthermore, FLIP-MCL requires additional projection layers for the contrastive loss ($L_{simCLR}$). Thus, FLIP-IT and FLIP-MCL have some auxiliary parameters, only during training. Moreover, since FLIP-MCL requires three forward passes through the image encoder (original + 2 transformed views), it involves more computations. Once the training is complete, the embeddings for the text prompts can be pre-computed and stored. Hence, all the auxiliary parameters (text encoder + proj) can be discarded and only the image encoder is required for inference. Therefore, our inference time is similar to the baseline method \cite{huang2022adaptive}, while our approach significantly improves the generalization to unseen domains.\\

\section{Robustness to Unseen Spoof Type}
\label{sec:additional_experiment_unseen_spoof_type}
To understand the robustness of the proposed FLIP-MCL method to unseen spoof types, we design an experiment to evaluate its performance, where the training and testing spoof types are completely different. We present the results in Table \ref{tab:suppl_unseen_spoof_type}. Each dataset in Protocol 1 (M, C, I, O) contains real, print attack, and replay attack samples. We aggregate the samples of real, print, and replay from all 4 datasets and split each group into a train-test split of 80\%-20\%. For the \textit{Replay} experiment, we train only on real and print samples and test on unseen replay samples. Similarly, we perform the \textit{Print} experiment by training only on real and replay samples and testing on unseen print samples. We observe that for both the unseen testing scenarios (\textit{Replay} \& \textit{Print}) the proposed FLIP-MCL method comfortably outperforms the baseline ViT thus demonstrating its generalizability. This validates the idea that aligning images to text descriptions can also handle unseen spoof types. \\
\input{tables/supplementary/suppl_unseen_spoof_type}

\end{document}

%% file: tables/table_mcio.tex
\begin{table*}[!t]
\centering \small
\caption{Evaluation of cross-domain performance in Protocol 1, between MSU-MFSD (\textbf{M}), CASIA-MFSD (\textbf{C}), Replay Attack (\textbf{I}) and OULU-NPU (O). We run each experiment 5 times under different seeds and report the mean HTER, AUC, and TPR@FPR=1\%. }
\label{tab:cross_db_mcio}
\setlength{\tabcolsep}{3pt}
      \scalebox{0.78}[0.78]{
	\begin{tabular}{llccccccccccccccc|cc}
	\multicolumn{17}{c}{} \\ 
        \toprule
	  & \multirow{3}{*}{\bf ~~Method~} & \multicolumn{3}{c}{\textbf{OCI} $\rightarrow$ \textbf{M}}  && \multicolumn{3}{c}{\textbf{OMI} $\rightarrow$ \textbf{C}}  &&  \multicolumn{3}{c}{\textbf{OCM} $\rightarrow$ \textbf{I}} && \multicolumn{3}{c}{\textbf{ICM} $\rightarrow$ \textbf{O}} && \textbf{Avg.} \\
	 \cmidrule{3-5} \cmidrule{7-9} \cmidrule{11-13} \cmidrule{15-17} \cmidrule{18-19} 
	 && \multirow{2}{*}{~HTER~} & \multirow{2}{*}{~AUC~} & TPR@ && \multirow{2}{*}{~HTER~} & \multirow{2}{*}{~AUC~} & TPR@ && \multirow{2}{*}{~HTER~} & \multirow{2}{*}{~AUC~} & TPR@ && \multirow{2}{*}{~HTER~} & \multirow{2}{*}{~AUC~} & TPR@ && \multirow{2}{*}{~HTER~} \\
	 &&&& ~FPR=$1\%$~ &&&& ~FPR=$1\%$~ &&&& ~FPR=$1\%$~ &&&& ~FPR=$1\%$~ \\ 
    \midrule
    \multirow{13}{*}{0-shot} 
    &~~MADDG (CVPR' 19) ~\cite{shao2019multi}        & 17.69 & 88.06 & -- && 24.50 & 84.51 & -- && 22.19 & 84.99 & -- && 27.98 & 80.02 & -- && 23.09\\
    &~~MDDR (CVPR' 20) ~\cite{wang2020cross}    & 17.02 & 90.10 & -- && 19.68 & 87.43 & -- && 20.87 & 86.72 & -- && 25.02 & 81.47 & -- && 20.64\\
    &~~NAS-FAS (TPAMI' 20) ~\cite{yu2020fas}         & 16.85 & 90.42 & -- && 15.21 & 92.64 & -- && 11.63 & 96.98 & -- && 13.16 & 94.18 & -- && 14.21\\
    &~~RFMeta (AAAI' 20) ~\cite{shao2020regularized}       & 13.89 & 93.98 & -- && 20.27 & 88.16 & -- && 17.30 & 90.48 & -- && 16.45 & 91.16 & -- && 16.97\\
    &~~$D^2$AM (AAAI' 21) ~\cite{chen2021generalizable}              & 12.70 & 95.66 & -- && 20.98 & 85.58 & -- && 15.43 & 91.22 & -- && 15.27 & 90.87 & -- && 16.09\\
    &~~DRDG (IJCAI' 21) ~\cite{liu2021dual}           & 12.43 & 95.81 & -- && 19.05 & 88.79 & -- && 15.56 & 91.79 & -- && 15.63 & 91.75 & -- && 15.66\\
    &~~Self-DA (AAAI' 21) ~\cite{wang2021self}              & 15.40 & 91.80 & -- && 24.50 & 84.40 & -- && 15.60 & 90.10 & -- && 23.10 & 84.30 & -- && 19.65\\
    &~~ANRL (ACM MM' 21) ~\cite{liu2021adaptive}                   & 10.83 & 96.75 & -- && 17.85 & 89.26 & -- && 16.03 & 91.04 & -- && 15.67 & 91.90 & -- && 15.09\\
    &~~FGHV (AAAI' 21) ~\cite{liu2022feature}                 & 9.17  & 96.92 & -- && 12.47 & 93.47 & -- && 16.29 & 90.11 & -- && 13.58 & 93.55 & -- && 12.87\\
    &~~SSDG-R (CVPR' 20) ~\cite{jia2020single}       & 7.38  & 97.17 & -- && 10.44 & 95.94 & -- && 11.71 & 96.59 & -- && 15.61 & 91.54 & -- && 11.28\\
    &~~SSAN-R (CVPR' 22) ~\cite{wang2022domain}       & 6.67  & 98.75 & -- && 10.00 & 96.67 & -- && 8.88  & 96.79 & -- && 13.72 & 93.63 & -- && 9.80\\
    &~~PatchNet (CVPR' 22) ~\cite{wang2022patchnet}       &  7.10 & 98.46 & -- && 11.33 & 94.58 & -- && 13.40 & 95.67 & -- && 11.82 & 95.07 & -- && 10.90\\
    &~~GDA (ECCV' 22) ~\cite{zhou2022generative}       & 9.20  & 98.00 & -- && 12.20 & 93.00 & -- && 10.00 & 96.00 & -- && 14.40 & 92.60 & -- && 11.45\\
     \midrule
     
     &~~DiVT-M (WACV' 23) \cite{liao2023domain}       & 2.86  & 99.14 & -- && 8.67  & 96.62 & -- && 3.71  & 99.29 & -- && 13.06 & 94.04 & -- && 7.07\\
     \multirow{-2}{*}{0-shot} &~~ViT (ECCV' 22) \cite{huang2022adaptive}         & \textbf{1.58} & \textbf{99.68} & \textbf{96.67} && 5.70 & 98.91 & 88.57 && 9.25 & 97.15 & 51.54 && 7.47 & 98.42 & 69.30 && 6.00 \\
     \hline
     &~~ViT (ECCV' 22) \cite{huang2022adaptive}         & 3.42 & 98.60 & 95.00 && 1.98 & 99.75 & 94.00 && 2.31 & 99.75 & 87.69 && 7.34 & 97.77 & 66.90 && 3.76 \\
     \multirow{-2}{*}{5-shot} &~~ViTAF* (ECCV' 22) \cite{huang2022adaptive}      & 2.92 & 99.62 & 91.66 && 1.40 & 99.92 & 98.57 && 1.64 & 99.64 & 91.53 && 5.39 & 98.67 & 76.05 && 3.31 \\
     \hline
     &~~FLIP-V                 & 3.79 & 99.31 & 87.99 && 1.27 & 99.75 & 95.85 && 4.71 & 98.80 & 75.84 && 4.15 & 98.76 & 66.47 && 3.48 \\
     &~~FLIP-IT                & 5.27 & 98.41 & 79.33 && \textbf{0.44} & \textbf{99.98} & 99.86 && \textbf{2.94} & \textbf{99.42} & \textbf{84.62} && 3.61 & 99.15 & 84.76 && 3.06 \\
     \rowcolor{green!12}
     \multirow{-3}{*}{0-shot} &~~FLIP-MCL                                & 4.95 & 98.11 & 74.67 && 0.54 & \textbf{99.98} & \textbf{100.00} && 4.25 & 99.07 & \textbf{84.62} && \textbf{2.31} & \textbf{99.63} & \textbf{92.28} && \textbf{3.01} \\

        \bottomrule
	\end{tabular}
}
\end{table*}

%% file: tables/table_wcs.tex
\begin{table*}[!t]
\centering
\caption{Evaluation of cross-domain performance in Protocol 2, between CASIA-SURF (\textbf{S}), CASIA-CeFA (\textbf{C}), and WMCA (\textbf{W}). We run each experiment 5 times under different seeds and report the mean HTER, AUC, and TPR@FPR=1\%  }
\label{tab:cross_db_wcs}
    \resizebox{0.95\textwidth}{!}
        {
	\begin{tabular}{llccccccccccc|cc}
	\multicolumn{13}{c}{} \\ 
        \toprule
	 & \multirow{3}{*}{\bf ~~Method~} & \multicolumn{3}{c}{\textbf{CS} $\rightarrow$ \textbf{W}}  && \multicolumn{3}{c}{\textbf{SW} $\rightarrow$ \textbf{C}}  &&  \multicolumn{3}{c}{\textbf{CW} $\rightarrow$ \textbf{S}} &&\textbf{Avg.} \\
	 \cmidrule{3-5} \cmidrule{7-9} \cmidrule{11-13}  \cmidrule{14-15}
	 && \multirow{2}{*}{~HTER~} & \multirow{2}{*}{~AUC~} & TPR@ && \multirow{2}{*}{~HTER~} & \multirow{2}{*}{~AUC~} & TPR@ &&  \multirow{2}{*}{~HTER~} & \multirow{2}{*}{~AUC~} & TPR@ &&\multirow{2}{*}{~HTER~} \\
	 &&&& ~FPR=$1\%$~ &&&& ~FPR=$1\%$~ &&&& ~FPR=$1\%$~ \\ \midrule

       0-shot & ViT (ECCV' 22) ~\cite{huang2022adaptive} & 7.98 & 97.97 & 73.61 && 11.13 & 95.46 & 47.59 && 13.35 & 94.13 & 49.97 && 10.82 \\
       \midrule
       \multirow{2}{*}{5-shot}
       & ViT (ECCV' 22) ~\cite{huang2022adaptive} & 4.30 & 99.16 & 83.55 && 7.69 & 97.66 & 68.33 && 12.26 & 94.40 & 42.59 && 6.06 \\
       & ViTAF* (ECCV' 22) ~\cite{huang2022adaptive} & 2.91 & 99.71 & 92.65 && 6.00 & 98.55 & 78.56 && 11.60 & 95.03 & 60.12 && 5.12 \\
       \midrule
       \multirow{3}{*}{0-shot}
       & FLIP-V                         & 6.13 & 97.84 & 50.26  && 10.89 & 95.82 & 53.93 && 12.48 & 94.43 & 53.00 && 9.83 \\
       & FLIP-IT                        & 4.89 & 98.65 & 59.14 && 10.04 & 96.48 & 59.4 && 15.68 & 91.83 & 43.27 && 10.2 \\
       \rowcolor{green!12}
       & FLIP-MCL                      & \textbf{4.46} & \textbf{99.16} & \textbf{83.86} && \textbf{9.66} & \textbf{96.69} & \textbf{59.00} & & \textbf{11.71} & \textbf{95.21} & \textbf{57.98} &&  \textbf{8.61} \\


       \bottomrule

    \vspace{1em}
    \end{tabular}
}
\end{table*}

%% file: tables/table_one_to_one_mcio.tex
\begin{table*}[!t]
    \centering \small
    \caption{Evaluation of cross-domain performance in Protocol 3, for all the 12 different combinations between MSU-MFSD (\textbf{M}), CASIA-MFSD (\textbf{C}), Replay Attack (\textbf{I}) and OULU-NPU (O). We run each experiment 5 times under different seeds and report the mean HTER.  }
    \label{tab:result_da}
   \setlength{\tabcolsep}{3pt}
      \scalebox{0.85}[0.85]{
    \begin{tabular}{llcccccccccccc|c}
    \multicolumn{14}{c}{} \\
    \toprule
    &
    \textbf{Method} & 
    \textbf{C} $\rightarrow$ \textbf{I} & 
    \textbf{C} $\rightarrow$ \textbf{M} & 
    \textbf{C} $\rightarrow$ \textbf{O} &
    \textbf{I} $\rightarrow$ \textbf{C} & 
    \textbf{I} $\rightarrow$ \textbf{M} & 
    \textbf{I} $\rightarrow$ \textbf{O} &
    \textbf{M} $\rightarrow$ \textbf{C} & 
    \textbf{M} $\rightarrow$ \textbf{I} & 
    \textbf{M} $\rightarrow$ \textbf{O} &
    \textbf{O} $\rightarrow$ \textbf{C} & 
    \textbf{O} $\rightarrow$ \textbf{I} & 
    \textbf{O} $\rightarrow$ \textbf{M} &
    \textbf{Avg.}\\
    
    \midrule
    
    \multirow{10}{*}{0-shot} 
    & ADDA (CVPR' 17) ~\cite{tzeng2017adversarial}           &41.8&36.6&-&49.8&35.1&-&39.0&35.2&-&-&-&-&39.6\\
    & DRCN (ECCV' 16) ~\cite{ghifary2016deep}                &44.4&27.6&-&48.9&42.0&-&28.9&36.8&-&-&-&-&38.1\\
    & DupGAN (CVPR' 18) ~\cite{hu2018duplex}                 &42.4&33.4&-&46.5&36.2&-&27.1&35.4&-&-&-&-&36.8\\
    & KSA (TIFS' 18) ~\cite{li2018unsupervised}                         &39.3&15.1&-&12.3&33.3&-&9.1&34.9&-&-&-&-&24.0\\
    & DR-UDA (TIFS' 20) ~\cite{wang2020unsupervised}         &15.6&9.0&28.7&34.2&29.0&38.5&16.8&3.0&30.2&19.5&25.4&27.4&23.1\\
    & MDDR (CVPR' 20) ~\cite{wang2020cross}                  &26.1&20.2&24.7&39.2&23.2&33.6&34.3&8.7&31.7&21.8&27.6&22.0&26.1\\
    & ADA (ICB' 19) ~\cite{wang2019improving}                &17.5&9.3&29.1&41.5&30.5&39.6&17.7&5.1&31.2&19.8&26.8&31.5&25.0\\
    & USDAN-Un (PR' 21) ~\cite{jia2021unified}               &16.0&9.2&-&30.2&25.8&-&13.3&3.4&-&-&-&-&16.3\\
    & GDA (ECCV' 22) ~\cite{zhou2022generative}              & 15.10 & \textbf{5.8} & - & 29.7 & 20.8 & - & 12.2 & 2.5 & - & - & - & - & 14.4 \\
    
    & CDFTN-L (AAAI' 23) ~\cite{yue2022cyclically}           &\textbf{1.7}& 8.1 &29.9&11.9& 9.6 &29.9& 8.8 &\textbf{1.3}&25.6 & 19.1&5.8& 6.3 & 13.2\\

    \midrule
    & FLIP-V            & 15.08 & 13.73 & 12.34 &  4.30  & 9.68  & 7.87  & 0.56 & 3.96  & 4.79 & 2.09 & 5.01  & 6.00 & 7.12 \\
    & FLIP-IT           & 12.33 & 15.18 & 7.98  & 1.12  & 8.37  & 6.98  & \textbf{0.19} & 5.21  & 4.96 & \textbf{0.16} & \textbf{4.27}  & \textbf{5.63} & 6.03 \\
    \rowcolor{green!12}
    \multirow{-3}{*}{0-shot}
    & FLIP-MCL         & 10.57 & 7.15  & \textbf{3.91} & \textbf{0.68} & \textbf{7.22} & \textbf{4.22} & \textbf{0.19} & 5.88 & \textbf{3.95} & 0.19 & 5.69 & 8.40 & \textbf{4.84} \\    

    
    \bottomrule
    \end{tabular}} \\

\end{table*}

%% file: tables/table_ablation_pretraining_compare.tex
\begin{table}[t]
\centering \small
\caption{Comparing different ViT initialization methods for FAS. We use each initialization method with their default parameters and show the results for \textbf{Protocol 1}.}
\label{tab:ablation_initialization}
\setlength{\tabcolsep}{3pt}
      \scalebox{0.65}[0.65]{
    \begin{tabular}{l|ccc|ccc|ccc|ccc|c}
    
    \multicolumn{9}{c}{}\\
    \toprule
    \multirow{2}{*}{\textbf{Method}} & \multicolumn{3}{c}{\textbf{OCI} $\rightarrow$ \textbf{M}} & \multicolumn{3}{c}{\textbf{OMI} $\rightarrow$ \textbf{C}} & \multicolumn{3}{c}{\textbf{OCM} $\rightarrow$ \textbf{I}} & \multicolumn{3}{c}{\textbf{ICM} $\rightarrow$ \textbf{O}} & \textbf{Avg.} \\
    \cmidrule{2-14}
	 & {~HTER~} & {~AUC~} && {~HTER~} & {~AUC~} && {~HTER~} & {~AUC~} && {~HTER~} & {~AUC~} && {~HTER~} \\
        \midrule
    Scratch          & 18.32 & 87.36 && 40.05 & 61.13 && 19.22 & 88.15 && 29.72 & 73.66 && 25.86 \\
    BeIT \cite{bao2022beit}            & 4.73 & 98.46 && 7.86 & 96.62 && 13.51 & 92.42 && 15.19 & 91.95 && 8.70 \\
    ImageNet \cite{huang2022adaptive}        & \textbf{1.58} & \textbf{99.68} && 5.70 & 98.91 && 9.25 & 97.15 && 7.47 & 98.42 && 6.00\\
    \rowcolor{green!12}
    CLIP (FLIP-V)             & 3.79 & 99.31 && \textbf{1.27} & \textbf{99.75} && \textbf{4.71} & \textbf{98.80} && \textbf{4.15} & \textbf{98.76} && \textbf{3.48}\\
        \bottomrule
	\end{tabular}
    }
\end{table}

%% file: tables/table_ablation_effect_of_prompts.tex
\begin{table}[t]
\vspace{1.25em}
\centering \small
\caption{Impact of guidance with different text prompts (described in Table \ref{tab:ensemble_texts}). We use FLIP-IT and show the results for \textbf{Protocol 1}.}
\label{tab:ablation_prompts}
\setlength{\tabcolsep}{3pt}
      \scalebox{0.65}[0.65]{
    \begin{tabular}{c|ccc|ccc|ccc|ccc|c}
    
    \multicolumn{9}{c}{}\\
    \toprule
    \multirow{2}{*}{\textbf{Prompt}} & \multicolumn{3}{c}{\textbf{OCI} $\rightarrow$ \textbf{M}} & \multicolumn{3}{c}{\textbf{OMI} $\rightarrow$ \textbf{C}} & \multicolumn{3}{c}{\textbf{OCM} $\rightarrow$ \textbf{I}} & \multicolumn{3}{c}{\textbf{ICM} $\rightarrow$ \textbf{O}} & \textbf{Avg.} \\
    \cmidrule{2-14}
	 & {~HTER~} & {~AUC~} && {~HTER~} & {~AUC~} && {~HTER~} & {~AUC~} && {~HTER~} & {~AUC~} && {~HTER~} \\
        \midrule
    P1         & 6.00 & 98.17 && 0.54 & 99.97 && 3.60 & 99.19 && 3.47 & 99.24 && 3.40 \\
    P2         & 8.32 & 96.38 && 1.05 & 99.90 && 2.98 & 99.48 && 5.74 & 98.39 && 4.52 \\
    P3         & \textbf{4.68} & \textbf{98.43} && \textbf{0.21} & \textbf{99.99} && 4.30 & 99.06 && 4.07 & 99.02 && 3.31\\
    P4         & 5.78 & 97.91 && 0.65 & 99.93 && 3.72 & 99.21 && 3.54 & 99.28 && 3.42 \\
    P5         & 6.48 & 98.37 && 0.46 & 99.96 && \textbf{2.52} & \textbf{99.55} && 3.24 & 99.30 && 3.17\\
    P6         & 5.58 & 98.00 && 0.3 & 99.99 && 2.85 & 99.28 && \textbf{3.03} & \textbf{99.46} && \textbf{2.94} \\
    \rowcolor{green!12}
    Ensemble         & 5.27 & 98.41 && 0.44 & 99.98 && 2.94 & 99.42 && 3.61 & 99.15 && 3.06 \\
        \bottomrule
	\end{tabular}
    }
\end{table}

%% file: tables/supplementary/suppl_param_sen_analysis.tex
\begin{table}[!t]
\vspace{1.25em}
\centering \small
\caption{Average HTER performance under different loss weights for Protocol 1. $L_{mcl} = \alpha L_{ce} + \beta L_{simCLR} + \gamma L_{mse}$}
\label{tab:suppl_param_sens_analyis}
\setlength{\tabcolsep}{7pt}
      \scalebox{0.9}[0.9]{
      {
	\begin{tabular}{c|ccccc}
	\multicolumn{6}{c}{} \\ 
        \toprule
        (${\alpha}, {\beta}, {\gamma}$ ) & \textbf{(1,1,1)} & \textbf{(1,1,0)} & \textbf{(1,0,1)} & \textbf{(1,2,2)} & \textbf{(1,5,5)} \\
        \midrule
        {HTER} & 3.01 & 3.15 & 3.47 & 3.20 & 3.67 \\
        \bottomrule
	\end{tabular}}
}
\end{table}


%% file: tables/supplementary/suppl_mcio.tex
\begin{table*}[!t]
\centering \small
\caption{Extending evaluation of cross-domain performance in Protocol 1 from 0-shot to 5-shot. We evaluate between MSU-MFSD (\textbf{M}), CASIA-MFSD (\textbf{C}), Replay Attack (\textbf{I}), and OULU-NPU (O). We run each experiment 5 times under different seeds and report the mean HTER, AUC, and TPR@FPR=1\%, along with their standard deviation (shown in brackets under the mean scores).}
\label{tab:suppl_cross_db_mcio}
\setlength{\tabcolsep}{3pt}
      \scalebox{0.78}[0.78]{
	\begin{tabular}{llccccccccccccccc|cc}
	\multicolumn{17}{c}{} \\ 
        \toprule
	  & \multirow{3}{*}{\bf ~~Method~} & \multicolumn{3}{c}{\textbf{OCI} $\rightarrow$ \textbf{M}}  && \multicolumn{3}{c}{\textbf{OMI} $\rightarrow$ \textbf{C}}  &&  \multicolumn{3}{c}{\textbf{OCM} $\rightarrow$ \textbf{I}} && \multicolumn{3}{c}{\textbf{ICM} $\rightarrow$ \textbf{O}} && \textbf{Avg.} \\
	 \cmidrule{3-5} \cmidrule{7-9} \cmidrule{11-13} \cmidrule{15-17} \cmidrule{18-19} 
	 && \multirow{2}{*}{~HTER~} & \multirow{2}{*}{~AUC~} & TPR@ && \multirow{2}{*}{~HTER~} & \multirow{2}{*}{~AUC~} & TPR@ && \multirow{2}{*}{~HTER~} & \multirow{2}{*}{~AUC~} & TPR@ && \multirow{2}{*}{~HTER~} & \multirow{2}{*}{~AUC~} & TPR@ && \multirow{2}{*}{~HTER~} \\
	 &&&& ~FPR=$1\%$~ &&&& ~FPR=$1\%$~ &&&& ~FPR=$1\%$~ &&&& ~FPR=$1\%$~ \\ 
    \midrule
    \multirow{13}{*}{0-shot} 
    &~~MADDG (CVPR' 19) ~\cite{shao2019multi}        & 17.69 & 88.06 & -- && 24.50 & 84.51 & -- && 22.19 & 84.99 & -- && 27.98 & 80.02 & -- && 23.09\\
    &~~MDDR (CVPR' 20) ~\cite{wang2020cross}    & 17.02 & 90.10 & -- && 19.68 & 87.43 & -- && 20.87 & 86.72 & -- && 25.02 & 81.47 & -- && 20.64\\
    &~~NAS-FAS (TPAMI' 20) ~\cite{yu2020fas}         & 16.85 & 90.42 & -- && 15.21 & 92.64 & -- && 11.63 & 96.98 & -- && 13.16 & 94.18 & -- && 14.21\\
    &~~RFMeta (AAAI' 20) ~\cite{shao2020regularized}       & 13.89 & 93.98 & -- && 20.27 & 88.16 & -- && 17.30 & 90.48 & -- && 16.45 & 91.16 & -- && 16.97\\
    &~~$D^2$AM (AAAI' 21) ~\cite{chen2021generalizable}              & 12.70 & 95.66 & -- && 20.98 & 85.58 & -- && 15.43 & 91.22 & -- && 15.27 & 90.87 & -- && 16.09\\
    &~~DRDG (IJCAI' 21) ~\cite{liu2021dual}           & 12.43 & 95.81 & -- && 19.05 & 88.79 & -- && 15.56 & 91.79 & -- && 15.63 & 91.75 & -- && 15.66\\
    &~~Self-DA (AAAI' 21) ~\cite{wang2021self}              & 15.40 & 91.80 & -- && 24.50 & 84.40 & -- && 15.60 & 90.10 & -- && 23.10 & 84.30 & -- && 19.65\\
    &~~ANRL (ACM MM' 21) ~\cite{liu2021adaptive}                   & 10.83 & 96.75 & -- && 17.85 & 89.26 & -- && 16.03 & 91.04 & -- && 15.67 & 91.90 & -- && 15.09\\
    &~~FGHV (AAAI' 21) ~\cite{liu2022feature}                 & 9.17  & 96.92 & -- && 12.47 & 93.47 & -- && 16.29 & 90.11 & -- && 13.58 & 93.55 & -- && 12.87\\
    &~~SSDG-R (CVPR' 20) ~\cite{jia2020single}       & 7.38  & 97.17 & -- && 10.44 & 95.94 & -- && 11.71 & 96.59 & -- && 15.61 & 91.54 & -- && 11.28\\
    &~~SSAN-R (CVPR' 22) ~\cite{wang2022domain}       & 6.67  & 98.75 & -- && 10.00 & 96.67 & -- && 8.88  & 96.79 & -- && 13.72 & 93.63 & -- && 9.80\\
    &~~PatchNet (CVPR' 22) ~\cite{wang2022patchnet}       &  7.10 & 98.46 & -- && 11.33 & 94.58 & -- && 13.40 & 95.67 & -- && 11.82 & 95.07 & -- && 10.90\\
    &~~GDA (ECCV' 22) ~\cite{zhou2022generative}       & 9.20  & 98.00 & -- && 12.20 & 93.00 & -- && 10.00 & 96.00 & -- && 14.40 & 92.60 & -- && 11.45\\
     \midrule
     
     &~~DiVT-M (WACV' 23) \cite{liao2023domain}       & 2.86  & 99.14 & -- && 8.67  & 96.62 & -- && 3.71  & 99.29 & -- && 13.06 & 94.04 & -- && 7.07\\
     \multirow{-2}{*}{0-shot} &~~ViT (ECCV' 22) \cite{huang2022adaptive}         & 1.58 & 99.68 & 96.67 && 5.70 & 98.91 & 88.57 && 9.25 & 97.15 & 51.54 && 7.47 & 98.42 & 69.30 && 6.00 \\
     \midrule
     &                 & 3.79 & 99.31 & 87.99 && 1.27 & 99.75 & 95.85 && 4.71 & 98.80 & 75.84 && 4.15 & 98.76 & 66.47 && 3.48 \\
     &~~\multirow{-2}{*}{FLIP-V}                  & (\textit{1.40}) & (\textit{0.31})  & (\textit{6.09})  && (\textit{0.85}) & (\textit{0.18})  & (\textit{5.53})  && (\textit{2.39}) & (\textit{0.85})  & (\textit{16.53}) && (\textit{0.56}) & (\textit{0.40})  & (\textit{14.97}) && (\textit{1.30}) \\
     &                  & 5.27 & 98.41 & 79.33 && {0.44} & {99.98} & 99.86 && {2.94} & {99.42} & {84.62} && 3.61 & 99.15 & 84.76 && 3.06 \\
     &~~\multirow{-2}{*}{FLIP-IT}                 & (\textit{1.3}) & (\textit{0.60}) & (\textit{10.93}) && (\textit{0.27}) & (\textit{0.02}) & (\textit{0.29}) && (\textit{1.3}) & (\textit{0.43}) & (\textit{15.14}) && (\textit{0.53}) & (\textit{0.19}) & (\textit{7.62}) && (\textit{0.80}) \\
     &                 & 4.95 & 98.11 & 74.67 && 0.54 & {99.98} & {100.00} && 4.25 & 99.07 & {84.62} && {2.31} & {99.63} & {92.28} && {3.01} \\
     \multirow{-6}{*}{0-shot}
     &~~\multirow{-2}{*}{FLIP-MCL}                 & (\textit{1.01}) & (\textit{0.50}) & (\textit{5.81}) && (\textit{0.22})  & (\textit{0.01}) & (\textit{0.00}) && (\textit{0.31}) & (\textit{0.17}) & (\textit{5.35}) && (\textit{0.46}) & (\textit{0.12}) & (\textit{3.37}) && (\textit{0.50})  \\
     
     \midrule
     \rowcolor{mygray}
     &~~ViT (ECCV' 22) \cite{huang2022adaptive}         & 3.42 & 98.60 & {95.00} && 1.98 & 99.75 & 94.00 && 2.31 & 99.75 & 87.69 && 7.34 & 97.77 & 66.90 && 3.76 \\
     \rowcolor{mygray}
     \multirow{-2}{*}{5-shot}
     &~~ViTAF* (ECCV' 22) \cite{huang2022adaptive}      & 2.92 & 99.62 & 91.66 && 1.40 & 99.92 & 98.57 && 1.64 & 99.64 & 91.53 && 5.39 & 98.67 & 76.05 && 3.31 \\
     \midrule
     
     \rowcolor{green!6}
     &                                   & {1.89} & {99.67} & {94.66} && 1.01 & 99.84 & 96.56 && 1.68 & 99.47 & 75.53 && 2.27 & 99.62 & 93.23 && {1.72} \\ 
     \rowcolor{green!6}
     &~~\multirow{-2}{*}{FLIP-V}                 & (\textit{0.63}) & (\textit{0.13}) & (\textit{3.39}) && (\textit{0.67}) & (\textit{0.14}) & (\textit{5.48}) && (\textit{0.69}) & (\textit{0.38}) & (\textit{22.07}) && (\textit{0.65}) & (\textit{0.15}) & (\textit{5.42}) && (\textit{0.66}) \\

     \rowcolor{green!6}
     &                                   & 2.63 & 99.55 & 93.00 && {0.46} & {99.97} & {99.86} && {1.18} & {99.83} & {96.15} && 3.07 & 99.30 & 83.15 && 1.83 \\
     \rowcolor{green!6}
     &~~\multirow{-2}{*}{FLIP-IT}                 & (\textit{0.78}) & (\textit{0.10}) & (\textit{3.71}) && (\textit{0.29}) & (\textit{0.02}) & (\textit{0.29}) && (\textit{0.26}) & (\textit{0.06}) & (\textit{1.95}) && (\textit{0.55}) & (\textit{0.06}) & (\textit{3.00}) && (\textit{0.47}) \\

     \rowcolor{green!6}
     &                                   & 3.42 & 99.34 & 82.67 && 0.63 & {99.98} & {100.00} && 1.52 & {99.86} & {97.23} && {1.54} &{99.81} & {96.37} && {1.77} \\
     \rowcolor{green!6}
     \multirow{-6}{*}{5-shot}
     &~~\multirow{-2}{*}{FLIP-MCL}                 & (\textit{0.16}) & (\textit{0.13}) & (\textit{7.35}) && (\textit{0.06}) & (\textit{0.01}) & (\textit{0.00}) && (\textit{0.09}) & (\textit{0.06}) & (\textit{1.04}) && (\textit{0.30}) & (\textit{0.06}) & (\textit{2.22}) && (\textit{0.15}) \\

        \bottomrule
	\end{tabular}
}
\end{table*}

%% file: tables/supplementary/suppl_wcs.tex
\begin{table*}[!t]
\vspace{5em}
\centering \small
\caption{Extending evaluation of cross-domain performance in Protocol 2 from 0-shot to 5-shot. We evaluate CASIA-SURF (\textbf{S}), CASIA-CeFA (\textbf{C}), and WMCA (\textbf{W}). We run each experiment 5 times under different seeds and report the mean HTER, AUC, and TPR@FPR=1\%, along with their standard deviation (shown in brackets under the mean scores).}
\label{tab:suppl_cross_db_wcs}
    \resizebox{0.95\textwidth}{!}
        {
	\begin{tabular}{llccccccccccc|cc}
	\multicolumn{13}{c}{} \\ 
        \toprule
	 & \multirow{3}{*}{\bf ~~Method~} & \multicolumn{3}{c}{\textbf{CS} $\rightarrow$ \textbf{W}}  && \multicolumn{3}{c}{\textbf{SW} $\rightarrow$ \textbf{C}}  &&  \multicolumn{3}{c}{\textbf{CW} $\rightarrow$ \textbf{S}} &&\textbf{Avg.} \\
	 \cmidrule{3-5} \cmidrule{7-9} \cmidrule{11-13}  \cmidrule{14-15}
	 && \multirow{2}{*}{~HTER~} & \multirow{2}{*}{~AUC~} & TPR@ && \multirow{2}{*}{~HTER~} & \multirow{2}{*}{~AUC~} & TPR@ &&  \multirow{2}{*}{~HTER~} & \multirow{2}{*}{~AUC~} & TPR@ &&\multirow{2}{*}{~HTER~} \\
	 &&&& ~FPR=$1\%$~ &&&& ~FPR=$1\%$~ &&&& ~FPR=$1\%$~ \\ \midrule

       0-shot & ViT (ECCV' 22) ~\cite{huang2022adaptive} & 7.98 & 97.97 & 73.61 && 11.13 & 95.46 & 47.59 && 13.35 & 94.13 & 49.97 && 10.82 \\
       \midrule
       &                                         & 6.13 & 97.84 & 50.26  && 10.89 & 95.82 & 53.93 && 12.48 & 94.43 & 53.00 && 9.83 \\
       & \multirow{-2}{*}{FLIP-V}                          & (\textit{2.24}) & (\textit{1.54}) & (\textit{25.05}) && (\textit{1.93}) & (\textit{1.27}) & (\textit{8.27}) && (\textit{1.26}) & (\textit{0.97}) & (\textit{6.27}) && (\textit{1.35}) \\
       &                                         & 4.89 & 98.65 & 59.14 && 10.04 & 96.48 & 59.4 && 15.68 & 91.83 & 43.27 && 10.2 \\
       & \multirow{-2}{*}{FLIP-IT}                         & (\textit{0.85}) & (\textit{0.48}) & (\textit{14.63}) && (\textit{0.46}) & (\textit{0.56}) & (\textit{5.48}) && (\textit{0.89}) & (\textit{0.75}) & (\textit{5.93}) && (\textit{0.55}) \\
       &                                         & 4.46 & {99.16} & {83.86} && {9.66} & {96.69} & {59.00} & & {11.71} & {95.21} & {57.98} &&  {8.61} \\
       \multirow{-6}{*}{0-shot}
       & \multirow{-2}{*}{FLIP-MCL}                         & (\textit{1.10}) & (\textit{0.31}) & (\textit{6.62}) && (\textit{0.50}) & (\textit{0.87}) & (\textit{8.87}) && (\textit{0.45}) & (\textit{0.38}) & (\textit{2.18}) && (\textit{0.51}) \\

       \midrule
       \rowcolor{mygray}
       & ViT (ECCV' 22) ~\cite{huang2022adaptive} & 4.30 & 99.16 & 83.55 && 7.69 & 97.66 & 68.33 && 12.26 & 94.40 & 42.59 && 6.06 \\
       \rowcolor{mygray}
       \multirow{-2}{*}{5-shot}
       & ViTAF* (ECCV' 22) ~\cite{huang2022adaptive} & 2.91 & 99.71 & 92.65 && 6.00 & 98.55 & 78.56 && 11.60 & 95.03 & 60.12 && 5.12 \\
       \midrule

       \rowcolor{green!6}
       &                                     & 0.69 & 99.96 & 99.42 && 3.68 & 99.38 & 85.87 && 7.44 & 97.62 & 76.11 && 2.95 \\
       \rowcolor{green!6}
       & \multirow{-2}{*}{FLIP-V}                      & (\textit{0.28}) & (\textit{0.05}) & (\textit{0.52}) && (\textit{1.32}) & (\textit{0.44}) & (\textit{7.05}) && (\textit{0.36}) & (\textit{0.27}) & (\textit{0.59}) && (\textit{0.49}) \\

       \rowcolor{green!6}
       &                                     & 0.80 & 99.96 & 98.67 && 3.19 & 99.44 & 88.80 && 7.63 & 97.42 & 71.6 && 2.90 \\
       \rowcolor{green!6}
       & \multirow{-2}{*}{FLIP-IT}                     & (\textit{0.44}) & (\textit{0.05}) & (\textit{1.40}) && (\textit{0.16}) & (\textit{0.11}) & (\textit{4.44}) && (\textit{0.60}) & (\textit{0.38}) & (\textit{3.49}) && (\textit{0.30}) \\

       \rowcolor{green!6}
       &                                     & 2.43 & 99.67 & 95.16 && 2.13 & 99.74 & 93.93 && 6.2 & 98.11 & 79.44 && 2.69 \\
       \rowcolor{green!6}
       \multirow{-6}{*}{5-shot} 
       & \multirow{-2}{*}{FLIP-MCL}                   & (\textit{0.78}) & (\textit{0.19}) & (\textit{2.4}) && (\textit{0.75}) & (\textit{0.13}) & (\textit{3.64}) && (\textit{0.53}) & (\textit{0.15}) & (\textit{1.29}) && (\textit{0.51}) \\
       
       \bottomrule
     
    \end{tabular}
}
\end{table*}

%% file: tables/supplementary/suppl_one_to_one.tex
\begin{table*}[!t]
    \centering \small
    \caption{Extending evaluation of cross-domain performance in Protocol 3 from 0-shot to 5-shot. We evaluate all the 12 different combinations between MSU-MFSD (\textbf{M}), CASIA-MFSD (\textbf{C}), Replay Attack (\textbf{I}), and OULU-NPU (O). We run each experiment 5 times under different seeds and report the mean HTER along with their standard deviation (shown in brackets under the mean scores).}
    \label{tab:suppl_result_da}
   \setlength{\tabcolsep}{3pt}
      \scalebox{0.85}[0.85]{
    \begin{tabular}{llcccccccccccc|c}
    \multicolumn{14}{c}{} \\
    \toprule
    &
    \textbf{Method} & 
    \textbf{C} $\rightarrow$ \textbf{I} & 
    \textbf{C} $\rightarrow$ \textbf{M} & 
    \textbf{C} $\rightarrow$ \textbf{O} &
    \textbf{I} $\rightarrow$ \textbf{C} & 
    \textbf{I} $\rightarrow$ \textbf{M} & 
    \textbf{I} $\rightarrow$ \textbf{O} &
    \textbf{M} $\rightarrow$ \textbf{C} & 
    \textbf{M} $\rightarrow$ \textbf{I} & 
    \textbf{M} $\rightarrow$ \textbf{O} &
    \textbf{O} $\rightarrow$ \textbf{C} & 
    \textbf{O} $\rightarrow$ \textbf{I} & 
    \textbf{O} $\rightarrow$ \textbf{M} &
    \textbf{Avg.}\\
    
    \midrule
    
    \multirow{10}{*}{0-shot} 
    & ADDA (CVPR' 17) ~\cite{tzeng2017adversarial}           &41.8&36.6&-&49.8&35.1&-&39.0&35.2&-&-&-&-&39.6\\
    & DRCN (ECCV' 16) ~\cite{ghifary2016deep}                &44.4&27.6&-&48.9&42.0&-&28.9&36.8&-&-&-&-&38.1\\
    & DupGAN (CVPR' 18) ~\cite{hu2018duplex}                 &42.4&33.4&-&46.5&36.2&-&27.1&35.4&-&-&-&-&36.8\\
    & KSA (TIFS' 18) ~\cite{li2018unsupervised}                         &39.3&15.1&-&12.3&33.3&-&9.1&34.9&-&-&-&-&24.0\\
    & DR-UDA (TIFS' 20) ~\cite{wang2020unsupervised}         &15.6&9.0&28.7&34.2&29.0&38.5&16.8&3.0&30.2&19.5&25.4&27.4&23.1\\
    & MDDR (CVPR' 20) ~\cite{wang2020cross}                  &26.1&20.2&24.7&39.2&23.2&33.6&34.3&8.7&31.7&21.8&27.6&22.0&26.1\\
    & ADA (ICB' 19) ~\cite{wang2019improving}                &17.5&9.3&29.1&41.5&30.5&39.6&17.7&5.1&31.2&19.8&26.8&31.5&25.0\\
    & USDAN-Un (PR' 21) ~\cite{jia2021unified}               &16.0&9.2&-&30.2&25.8&-&13.3&3.4&-&-&-&-&16.3\\
    & GDA (ECCV' 22) ~\cite{zhou2022generative}              & 15.10 & 5.8 & - & 29.7 & 20.8 & - & 12.2 & 2.5 & - & - & - & - & 14.4 \\
    & CDFTN-L (AAAI' 23) ~\cite{yue2022cyclically}           &{1.7}& 8.1 &29.9&11.9& 9.6 &29.9& 8.8 & {1.3}&25.6 & 19.1&5.8& 6.3 & 13.2\\

    \midrule
    &                           & 15.08 & 13.73 & 12.34 &  4.30  & 9.68  & 7.87  & 0.56 & 3.96  & 4.79 & 2.09 & 5.01  & 6.00 & 7.12 \\
    & \multirow{-2}{*}{FLIP-V}            & (\textit{4.60}) & (\textit{4.81}) & (\textit{4.41}) & (\textit{2.41})  & (\textit{1.62})  & (\textit{1.39})  & (\textit{0.46}) & (\textit{0.77})  & (\textit{0.98}) & (\textit{0.63}) & (\textit{1.41}) & (\textit{1.69}) & (\textit{2.10}) \\
    &                            & 12.33 & 15.18 & 7.98  & 1.12  & 8.37  & 6.98  & {0.19} & 5.21  & 4.96 & {0.16} & {4.27}  & {5.63} & 6.03 \\
    & \multirow{-2}{*}{FLIP-IT}            & (\textit{2.24}) & (\textit{2.40}) & (\textit{2.73}) & (\textit{0.30})  & (\textit{2.95}) & (\textit{1.14}) & (\textit{0.26}) & (\textit{2.57})  & (\textit{0.75}) & (\textit{0.22}) & (\textit{1.53})  & (\textit{1.61}) & (\textit{1.55}) \\
    
    &                            & 10.57 & {7.15}  & {3.91} & {0.68} & {7.22} & {4.22} & {0.19} & 5.88 & {3.95} & 0.19 & 5.69 & 8.40 & {4.84} \\
    \multirow{-6}{*}{0-shot}
    & \multirow{-2}{*}{FLIP-MCL}            & (\textit{2.94}) & (\textit{1.4}) & (\textit{0.47}) & (\textit{0.05}) & (\textit{2.15})  & (\textit{0.37})  & (\textit{0.20}) & (\textit{1.38}) & (\textit{0.42}) & (\textit{0.26}) & (\textit{1.42})  & (\textit{1.09}) & (\textit{1.01}) \\
    
    \midrule
    \rowcolor{mygray}
    &                               & 4.98 & 4.38 & 10.85 & 2.55 & 5.08 & 8.63 & 1.59 & 1.79 & 7.92 & 1.65 & 3.4 & 4.4 & 4.77 \\
    \rowcolor{mygray}
    & \multirow{-2}{*}{ViTAF*}     & (\textit{0.66}) & (\textit{0.80}) & (\textit{1.31}) & (\textit{0.34}) & (\textit{0.95}) & (\textit{0.97}) & (\textit{0.20}) & (\textit{0.13}) & (\textit{0.71}) & (\textit{0.36}) & (\textit{0.71}) & (\textit{0.73}) & (\textit{0.66}) \\
    \cmidrule{2-15}

    \rowcolor{green!6}
    &                               & 3.37 & {2.27} & {2.96} & 0.79 & 2.37 & 3.75 & 0.42 & 2.38 & {2.76} & 0.35 & {1.62} & 2.10 & {2.10} \\
    \rowcolor{green!6}
    & \multirow{-2}{*}{FLIP-V}     & (\textit{1.23}) & (\textit{1.19}) & (\textit{0.68}) & (\textit{0.26}) & (\textit{1.26}) & (\textit{0.92}) & (\textit{0.30}) & (\textit{0.34}) & (\textit{0.47}) & (\textit{0.29}) & (\textit{0.34}) & (\textit{0.68}) & (\textit{0.66}) \\

    \rowcolor{green!6}
    &                             & 4.11 & 5.22 & 4.20 & {0.42} & {2.22} & {3.20} & {0.40} & 2.31 & 3.21 & {0.16} & 2.45 & 3.78 & 2.64 \\
    \rowcolor{green!6}
    & \multirow{-2}{*}{FLIP-IT}     & (\textit{0.74}) & (\textit{0.57}) & (\textit{0.59}) & (\textit{0.25}) & (\textit{0.79}) & (\textit{0.33}) & (\textit{0.33}) & (\textit{0.65}) & (\textit{0.46}) & (\textit{0.22}) & (\textit{0.55}) & (\textit{0.73}) & (\textit{0.50}) \\

    \rowcolor{green!6}
    &                            & 4.18 & 5.27 & 2.48 & 0.65 & 3.68 & 2.56 & 0.19 & 1.74 & 2.43 & 0.23 & 2.58 & 4.10 & 2.51 \\
    \rowcolor{green!6}
    \multirow{-8}{*}{5-shot}
    & \multirow{-2}{*}{FLIP-MCL}     & (\textit{0.60}) & (\textit{0.53}) & (\textit{0.53}) & (\textit{0.06}) & (\textit{0.53}) & (\textit{0.42}) & (\textit{0.20}) & (\textit{0.29}) & (\textit{0.26}) & (\textit{0.23}) & (\textit{0.59}) & (\textit{1.25}) & (\textit{0.45}) \\
    
    \bottomrule
    \end{tabular}} \\

\end{table*}

%% file: tables/supplementary/suppl_computational_complexity.tex
\begin{table*}[!t]
\vspace{1.5em}
\centering \small
\caption{Computational complexity analysis for all the methods in the FLIP framework compared with the baseline methods.}
\label{tab:suppl_computaional_comp_params}
\vspace{-0.6em}
\scalebox{0.9}[0.9]{
    \begin{tabular}{l|cc|ccc|cc|cc}
    \multicolumn{6}{c}{}\\
        \toprule
            \multirow{3}{*}{\textbf{Method}} & \multicolumn{4}{c}{\textbf{Training}} && \multicolumn{2}{c}{\textbf{Inference}} && \multirow{2}{*}{\textbf{Inference Time}}\\
            & \multicolumn{2}{c}{\textbf{Image Encoder}} & \multicolumn{2}{c}{\textbf{Text Encoder + Proj}} &&&& \\
            & \textbf{Parameters} & \textbf{FLOPs} & \textbf{Parameters} & \textbf{FLOPs} && \multirow{-2}{*}{\textbf{Parameters}} & \multirow{-2}{*}{\textbf{FLOPs}} && \textbf{(seconds/frame)} \\
            \midrule
            ViT (ECCV' 22)      & 86.19M & 17.58G & - & - && 86.19M & 17.58G  && 0.007\\
            ViTAF* (ECCV' 22)   & 92.02M & 18.68G & - & -  && 92.02M & 18.68G && 0.020\\
            FLIP-V          & 86.58M  & 17.58G & - & - && 86.58M  & 17.58G && 0.013\\
            FLIP-IT         & 86.19M & 17.58G & 63.11M & 35.81G && 86.19M  & 17.58G && 0.010 \\
            FLIP-MCL        & 86.19M & 52.74G & 83.05M  & 35.86G && 86.19M  & 17.58G  && 0.010 \\
            \bottomrule
    \end{tabular}
}
\end{table*}

%% file: tables/supplementary/suppl_unseen_spoof_type.tex
\begin{table}[!t]
\centering \small

\caption{HTER performance on unseen spoof type at test time. \textit{Replay} denotes training on real+print samples and testing on unseen replay samples. \textit{Print} denotes training on real+replay samples and testing on unseen print samples.}
\label{tab:suppl_unseen_spoof_type}
\setlength{\tabcolsep}{6pt}
\scalebox{1}[1]{
 \begin{tabular}{l|cc}
	\multicolumn{3}{c}{} \\ 
        \toprule
        \textbf{Method} & \textbf{\textit{Replay}} & \textbf{\textit{Print}}\\
        \midrule
        {ViT (ECCV' 22) ~\cite{huang2022adaptive}} & 4.69 & 10.36 \\
        {FLIP-MCL} & 1.07 & 1.98\\
        \bottomrule
	\end{tabular}}

\end{table}